%% file: main.tex
\pdfoutput=1

\documentclass[11pt]{article}

\usepackage[final]{acl}

\usepackage{times}
\usepackage{latexsym}

\usepackage[T1]{fontenc}

\usepackage[utf8]{inputenc}

\usepackage{xcolor}
\usepackage{xspace}
\usepackage{colortbl}
\usepackage{multirow}
\usepackage{amsmath}
\usepackage{caption}
\usepackage{graphicx, subfig}
\usepackage{enumitem}
\usepackage{xspace}
\usepackage{makecell}
\usepackage{soul}

\usepackage{microtype}

\usepackage{inconsolata}

\usepackage{graphicx}

\title{DeepThink: Aligning Language Models with Domain-Specific User Intents}

\author{
Yang Li\textsuperscript{\rm 1,2}\footnotemark[1],\ \ 
Mingxuan Luo\textsuperscript{\rm 1},\ \ 
Yeyun Gong\textsuperscript{\rm 2}\footnotemark[2],\ \ 
Chen Lin\textsuperscript{\rm 1}\footnotemark[2],\ \ 
Jian Jiao\textsuperscript{\rm 2}
\\
\textbf{
Yi Liu\textsuperscript{\rm 2},
Kaili Huang\textsuperscript{\rm 2}
}
\\
\textsuperscript{\rm 1} 
Xiamen University 
\textsuperscript{\rm 2}
Microsoft
}

\begin{document}
\maketitle
\renewcommand{\thefootnote}{\fnsymbol{footnote}}
 \footnotetext[1]{Work done during an internship at Microsoft.}
 \footnotetext[2]{Corresponding author}

\newcommand{\ourmodel}{\texttt{DeepThink}\xspace}
\newcommand{\convaugment}{CDS\xspace}
\newcommand{\convaugmentfull}{Conversation-based Data Synthesis\xspace}
\newcommand{\convrefine}{CDR\xspace}
\newcommand{\convrefinefull}{Conversation-based Data Refinement\xspace}
\newcommand{\turn}{\mathcal{N}}
\newcommand{\maxiter}{\mathcal{M}}
\newcommand{\filterthreshold}
{r_{\theta}}
\newcommand{\question}{x}
\newcommand{\answer}{y}
\newcommand{\documents}{d}
\newcommand{\trainingloss}{\mathcal{L}}
\newcommand{\realuserQnum}{Q}
\newcommand{\Inquirer}{\textit{Inquirer}\xspace}
\newcommand{\Assistant}{\textit{Assistant}\xspace}
\newcommand{\pool}{\mathcal{P}}

\begin{abstract}
Supervised fine-tuning with synthesized instructions has been a common practice for adapting LLMs to domain-specific QA tasks. However, the synthesized instructions deviate from real user questions and expected answers. This study proposes a novel framework called DeepThink to generate high-quality instructions. DeepThink first generates a few seed questions to mimic actual user questions, simulates conversations to uncover the hidden user needs, and refines the answer by conversational contexts and the retrieved documents for more comprehensive answers. Experiments demonstrate that DeepThink achieves an average performance improvement of 7.92\% compared to a GPT-4-turbo+RAG-based assistant on the real user test set in the advertising domain across dimensions such as relevance, completeness, clarity, accuracy, and actionability.
\end{abstract}

\input{introduction}

\input{relatedworks}

\input{method}

\input{experiments}

\input{conclustion}
\bibliography{ref}
\input{appendix}

\end{document}

%% file: introduction.tex
\section{Introduction}
\label{sec:intro}
Large language models (LLMs) such as GPT-4~\cite{OpenAI:2023ktj} have achieved remarkable advancements in question-answering (QA) tasks. Commercial and open-source LLMs are primarily trained on general-domain data and perform less effectively in vertical domains such as healthcare, finance and advertising~\cite{goyal2024healai,xu2024mental,wu2023bloomberggpt}. Supervised Fine-Tuning~\cite{sanh2022multitask,wang2024instruct,Wang2022SelfInstructAL} (SFT) has been widely adopted to optimize the LLM's parameters on a curated set of instructions or task examples to enhance LLMs' ability to answer domain-specific questions. 

Due to the high cost of collecting instruction data, recent SFT methods generate synthetic data. They typically start with a few \textit{seed instructions}, either constructed manually~\cite{ouyang2022training} or generated by LLMs from available documents~\cite{wang2024instruct}. The seed questions are then expanded~\cite{Wang2022SelfInstructAL} or evolved~\cite{Xu2023WizardLMEL} to provide greater complexity and diversity. However, the synthesized instructions deviate from user questions and expected answers.

\begin{figure}[!t]
\centering
\includegraphics[width=1\columnwidth]{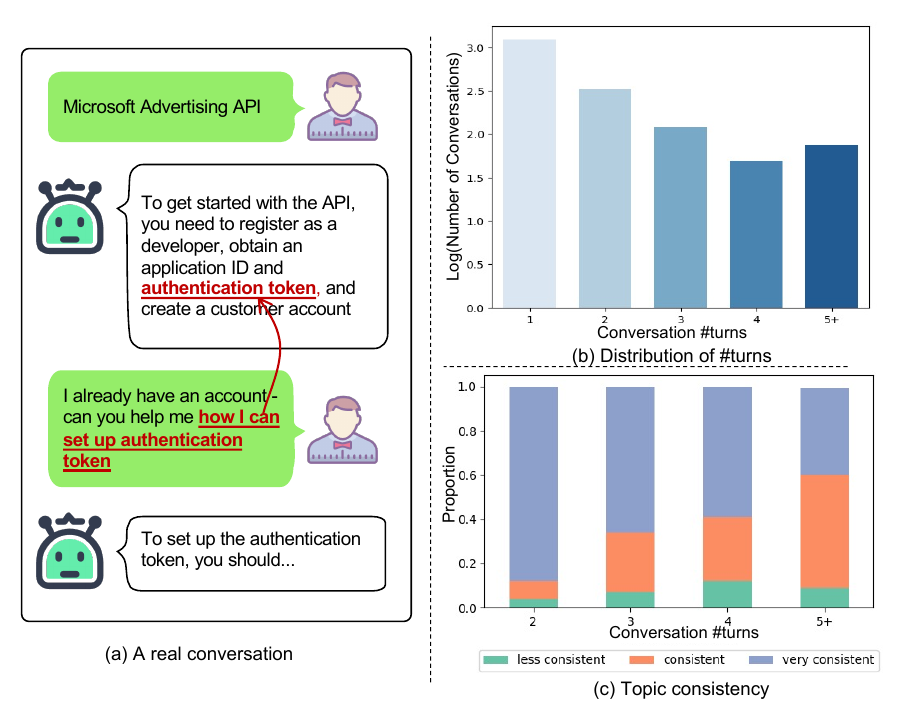} 
\caption{Three phenomena on real-world advertising platforms}
\label{fig:discoverys}
\end{figure}

We now characterize the user demands of domain-specific QA by analyzing an advertising platform. As illustrated in Figure~\ref{fig:discoverys}(a), users usually start with a brief question and ask for more details mentioned in the assistant's responses. Consequently, multi-turn dialogues constitute a substantial portion (Figure~\ref{fig:discoverys}(b)), and most conversations, no matter how many turns, focus on one topic theme  (Figure~\ref{fig:discoverys}(c)). \footnote{The platform currently uses GPT-4-turbo with Retrieval Augmented Generation (RAG) as an intelligent assistant. We prompt GPT-4-turbo to analyze whether the dialogue is closely centered around a single topic theme and assign a topic consistency score. The scores range from 1 to 5, where a score of 3 or below indicates a lower level of thematic consistency, and a score of 5 represents very high thematic consistency.}  

The above observations reveal a critical challenge of QA in vertical domains: Questions are vague and incomplete and do not reflect the user's hidden interests. Because specialized knowledge is required in vertical domains, users do not always possess such expertise, and they are incapable of asking concise and accurate questions. LLMs fine-tuned by conventional instructions (e.g., synthetic questions that mismatch actual user behavior patterns and answers that fall short of user expectations) tend to provide broad responses, which will increase the cost of consultation and harm user experience. The question naturally arises. \textit{
Could we use high-quality synthetic instructions to fine-tune the LLM to capture the user's hidden interests and give a precise and relevant answer?} 

Figure~\ref{fig:discoverys}(a) also demonstrates that authentic user interests are gradually exposed through conversations and satisfying answers are obtained by continuously expanding on a question and delving into technical details. Inspired by this insight, we propose a novel supervised fine-tuning method named \ourmodel. \ourmodel first generates a few seed questions guided by actual user questions. Then, to simulate human conversations, \ourmodel designs a dual-role (i.e., the inquirer and the assistant) framework to generate dialogues on the seed questions. An evaluator assesses each answer based on the dialogue context and provides revision suggestions for a refiner to enhance the quality of the answers. Finally, the questions in each turn of the simulated dialogues and the corresponding refined answers are employed for supervised fine-tuning.

\begin{figure}[htbp]
\centering
\includegraphics[width=\columnwidth]{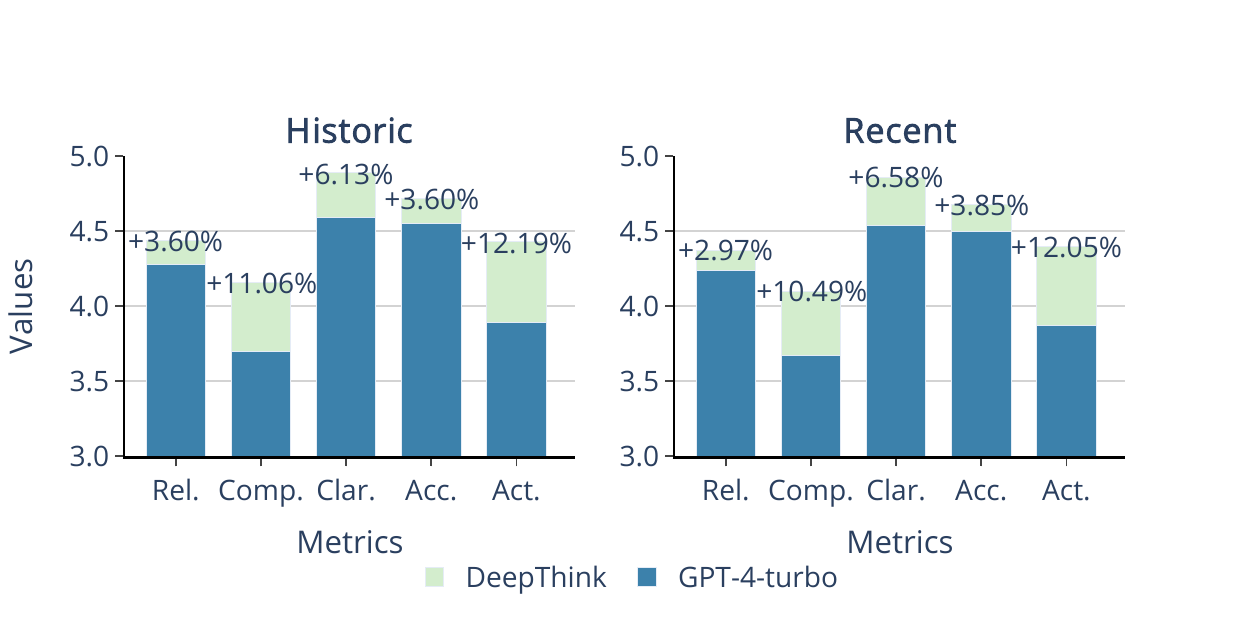} 
\caption{Performance comparison of DeepThink and GPT-4-turbo across five evaluation dimensions over different time spans ( "Historic," and "Recent."). DeepThink performs better than GPT-4-turbo in relevance, completeness, clarity, accuracy, and actionability.}
\label{fig:comp}
\end{figure}

We evaluate the performance of \ourmodel on an online advertising platform. As shown in Figure~\ref{fig:comp}, \ourmodel surpasses commercial LLM, i.e., GPT-4-turbo+RAG and achieves improvements of 3.43\%, 12.09\%, 6.69\%, 3.74\%, and 13.69\% regarding relevance, completeness, clarity, accuracy, and actionability, respectively.

In summary, the main contribution of \ourmodel is threefold. (1) \ourmodel presents a novel instruction synthesis method by simulating real-world user queries and follow-up conversations for supervised fine-tuning. (2) \ourmodel refines the synthesized answers based on the conversation contexts to ensure that the LLMs can generate more comprehensive answers and address user interests in vertical domains. (3) A large-scale evaluation on an advertising platform of real user questions verifies the effectiveness of \ourmodel.

%% file: relatedworks.tex
\section{Related Works}


\subsection{Instruction Data Synthesis}
To address the issue of limited training samples in specific domains, various works have proposed using additional data, such as manual annotation~\cite{Zhao2024WildChat1C,zheng2023lmsys} and automatic generation by LLMs~\cite{Mekala2022LeveragingQD, Wang2021TowardsZL, Wang2022SelfInstructAL, Xu2023WizardLMEL}. However, manual annotation is expensive~\cite{honovich-etal-2023-unnatural}, and iterative generation by LLMs frequently introduces the risk of hallucinations.

Our work falls into the category of automatic generation by LLMs. However, our work differs from previous approaches in two main aspects. (1) We synthesize instructions by simulating conversations closer to real-world scenarios. (2) We adopt several techniques to improve the quality of synthesized instruction. We integrate Retrieval-Augmented Generation (RAG) to mitigate hallucination in conversation-based synthesis. We apply a Conversation-based Data Refiner for filtering, ensuring topic consistency and data authenticity.
\subsection{Retrieval-Augmented Generation}
Retrieval augmentation has become a standard solution to address hallucinations in LLMs by introducing external knowledge to compensate for factual shortcomings~\cite{Asai2023SelfRAGLT,ma2023query, Izacard2021UnsupervisedDI, Ram2023InContextRL}.
Early Retrieval Augmentation efforts focus primarily on the retriever itself, where both the neural retriever and generator are typically trainable Pretrained Language Models (PrLMs), such as BERT ~\cite{Devlin2019BERTPO} or BART ~\cite{Lewis2019BARTDS}. In contrast, modern Retrieval Augmentation applied to LLMs emphasizes determining when and how to retrieve relevant information~\cite{fatehkia2024t, Asai2023SelfRAGLT, Xu2024LargeLM}. For example, Self-RAG enables on-demand retrieval and generates more accurate, fact-based text through fine-grained self-reflection~\cite{Asai2023SelfRAGLT}. 

Our approach uses RAG throughout the data synthesis, SFT, and inference stages. This not only improves the authenticity of the synthesized data but also helps the LLM learn how to effectively utilize the retrieved knowledge during the SFT stage. In contrast, previous research only used RAG during the inference stage, relying heavily on the LLM's ability to discern the retrieved knowledge. This can lead to insufficient utilization of relevant knowledge, especially when dealing with domain knowledge that was not included in the pretraining process.

%% file: method.tex
\section{Approach}
\begin{figure*}[htbp]
\centering
\includegraphics[width=0.98\linewidth]{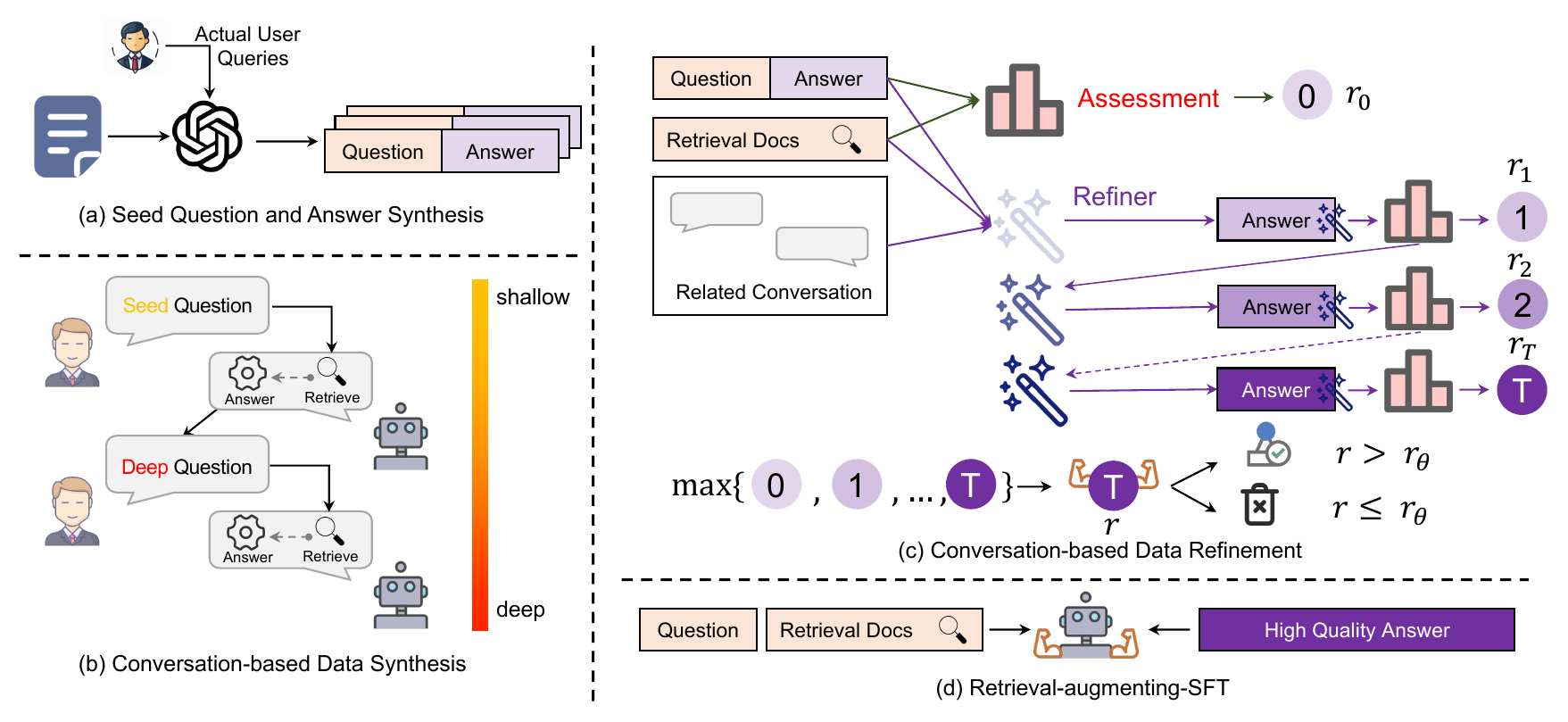}
    \caption{The framework of \ourmodel}
    \label{fig:framework}
\end{figure*}

As illustrated in Figure~\ref{fig:framework}, \ourmodel consists of four key stages: (1) Seed Question and Answer Synthesis, (2) Conversation-based Data Synthesis, (3) Conversation-based Data Refinement, and (4) Retrieval-augmenting-SFT.

\subsection{Seed Question and Answer Synthesis}
We leverage GPT-4-turbo to synthesize instructions. Existing studies such as SELF-QA~\cite{zhang2023self} utilize LLMs to extract questions from documents, enabling the automatic generation of seed questions. Unfortunately, the questions generated through existing approaches exhibit significant linguistic style discrepancies compared with those found in genuine user-LLM interactions. To resolve this issue, we randomly sample a few authentic user questions and prompt GPT-4-turbo to generate queries that mirror the linguistic style and structure of these samples. This method ensures that the generated questions reflect the realistic language and format of the actual user queries. Details of the specific prompts used are in Appendix~\ref{sec:appPrompts}.

\subsection{Conversation-based Data Synthesis}
We implement a dual-role conversation framework based on GPT-4-turbo, where one role is designated as the \Inquirer and the other as the \Assistant. \ourmodel guides the two roles to emulate authentic and high-quality conversations. Specifically, in guiding the \Inquirer, \ourmodel instructs it to mimic the style of actual user inquiries by incorporating real user questions into \Inquirer's prompt. This stylistic imitation distinguishes from earlier methods~\cite{Wang2022SelfInstructAL} without instructing the LLM using actual user queries, resulting in dialogues that more closely reflect real-world conversational dynamics.

Previous methods rely on the inherent knowledge of LLMs to generate answers and often lead to hallucinations, especially in vertical domains where LLMs lack direct training data~\cite{abdullin2024synthetic,liu2024chatqa}. In generating the \Assistant's responses, \ourmodel incorporates a retrieval-augmented generation framework. By retrieving domain-relevant documents to ground responses, this approach mitigates the risk of hallucination and enhances the response's accuracy.

Furthermore, to maintain engagement and progressively deepen the dialogue, \ourmodel instructs the \Assistant to suggest follow-up questions based on topics that may interest the user. The \Inquirer then has the option to (1) choose from these suggestions, (2) generate a new question, or (3) respond with "No more questions" to end the conversation. This structured interaction ensures that the conversation flow remains natural. The interaction is enforced to end when exceeding a predefined maximal number of turns because long dialogues are likely to drift from the original topic. The prompt is shown in Figure~\ref{fig:assitantAgentPrompt}, and cases are shown in Appendix~\ref{sec:scrr}.

\subsection{Conversation-based Data Refinement}
The answers generated from the above procedures face several critical issues: (1) they merely provide superficial responses to user queries without capturing the underlying intent behind the questions, and (2) they fail to address ambiguous or unclear user queries, resulting in answers that do not align with the user's expectations. To mitigate these challenges, leveraging question-and-answer pairs from other turns in the conversation to supplement the current response presents a natural solution. However, this process is inherently complex, as the content from other turns may not always align perfectly with the current question, and irrelevant information should not be incorporated into the refinement. To address this, we propose an iterative answer refinement strategy based on the synthesized conversation. In each iteration of the refinement process, the refiner is prompted to refine the answer based on the conversational context, followed by an assessment phase where the refined content is evaluated and constructive feedback is generated. This feedback is then utilized as input for the subsequent iteration, guiding the refiner to improve the response further.

\textbf{Initialization.} Refinement focuses on enriching the current answer by incorporating relevant information from the conversation's other turns. Specifically, \ourmodel feeds synthesized questions, corresponding answers, and the related conversation context into GPT-4-turbo (Refiner). By designing specific prompts that guide the Refiner to mimic the linguistic style of real user inquiries, we ensure that the generated answers are both comprehensive and stylistically consistent with authentic user interactions. Additionally, to minimize irrelevant interference and prevent potential hallucinations, we retrieve documents closely aligned with the current question and include them in the input.

\textbf{Feedback-based Refinement}. As previous studies~\cite{zheng2024judging,mao2023gpteval} have demonstrated GPT-4-turbo's capability to emulate human evaluation preferences, we employ it as an effective assessor. GPT-4-turbo evaluates responses across five dimensions: relevance, completeness, clarity, accuracy, and actionability, providing an overall score and detailed feedback. This feedback is subsequently utilized as input for the refiner to further refine the response in the next iteration. This multi-faceted assessment allows for targeted refinements, ensuring that each aspect of the response aligns with user expectations and the conversational context. The iterative process continues for a maximum of rounds $T$.

\textbf{Instruction Update and Filtering.} We calculate the overall score $r_0$ of the original answer $a_0$ and put the original answer in the selection pool $\pool$. We also obtain the assessment score $r_t$ for the refined answer $a_t$ in each iteration, where $t\in[1,T]$. We put these answers in $\pool$. We select the best answer with the highest score in the pool that exceeds a predefined quality threshold, i.e., $r = r_{\arg\max_{0 \leq t\leq T}r_t,r_t> \filterthreshold}$.

\subsection{Retrieval-Augmented Supervised Fine-Tuning}
To effectively capture and utilize domain-specific knowledge, our proposed \ourmodel integrates Retrieval-Augmented Generation (RAG) within the Supervised Fine-Tuning (SFT) framework. Unlike traditional SFT approaches that rely solely on (question, answer) pairs generated from the model's inherent knowledge, \ourmodel leverages relevant external documents to enhance answer accuracy and context awareness.

\textbf{Motivation for RAG Integration.} In vertical domains such as advertising, healthcare, and finance, user queries often require precise and contextually rich responses that depend on up-to-date and domain-specific information. Traditional SFT methods fall short in these scenarios as they do not utilize external knowledge sources, limiting the model's ability to generate accurate and relevant answers based on the provided context.

\textbf{RAG-Augmenting-SFT.} Our approach involves incorporating retrieved documents into the supervised fine-tuning process. For each training instance, given a question $\question_i$, we first retrieve the most relevant documents $\documents_i$ from a curated knowledge base. The model then generates an answer $\answer_i$ conditioned on both the question and the retrieved documents. This results in a (question, document, answer) triplet that forms the basis of our fine-tuning data.

Formally, the training loss is defined as: \begin{equation} \trainingloss(\Phi) = -\sum_{(\question_i,\documents_i,\answer_i)}\log p(\answer_i|\question_i,\documents_i,\Phi), \end{equation} where $\Phi$ represents the LLM parameters, $\answer_i$ is the answer generated by the LLM, and $p(\cdot)$ is the likelihood of the answer given the question and documents.


%% file: experiments.tex
\section{Experiments}
\subsection{Experimental Setups}
\label{sec:setups}
\textbf{Dataset}. We consider the Advertising Domain in this paper. We have obtained relevant documentation from an advertising platform, consisting of a total of 7.5k text segments covering Setup and Basics, Manage Ads, Measuring Results, and Billing and Payments. For testing, we have collected real user questions from our advertising platform, spanning the period from March 27 to June 6. The dataset underwent a cleaning process, including deduplication and filtering out questions unrelated to advertising, resulting in a refined collection of 7,801 user questions. Subsequently, we order these questions chronologically based on the time they were posted and divide them into two subsets: (1) \textit{Historic}: consists of the earliest 90\% of data (i.e., 6,617 questions). This subset allows us to gauge the model's response quality over a broad range of topics, thus offering a comprehensive view of the model's overall capabilities. (2) \textit{Recent}: consists of the most recent 10\% of data (i.e., 1,184 questions), representing the latest user needs on the platform. It is used to assess the LLM's response quality for recent user activities. This subset helps detect any potential degradation in the model's performance over time, ensuring its continued reliability and alignment with user expectations.

\textbf{Imitation seed data.} Following prior work~\cite{zhang2023self}, we prompt GPT-4-turbo to generate at least one high-quality instruction for each text segment from the domain data. To approximate the distribution of real user inquiries, we randomly select 15 authentic user questions from our advertising platform to prompt GPT-4-turbo to generate seed questions. Our conversation-based Data synthesis also employs the same set of 15 real user questions throughout the process. We obtain approximately 5k seed data. Detailed prompts are provided in Figure~\ref{fig:docQAsPrompts}. 

\begin{table}[!t]
\caption{Statistics of Instruction Data Generated by Different Methods}
\label{tab:genQAsStatistic}
\resizebox{\linewidth}{!}{
\begin{tabular}{lllll}
\hline
              & \# Examples & \# Domain & \makecell[l]{Instruction \\Length} & \makecell[l]{Output\\ Length} \\
\hline
Seed data     & 5k          & advertising       & 10±3              & 87±20         \\
Self Instruct & 23k         & advertising       & 15±13             & 56±27         \\
Evol Instrcut & 15k         & advertising       & 18±8              & 74±24         \\
Magpie        & 300k        & general   & 12±6              & 377±76            \\
\ourmodel     & 12k         & advertising       & 18±6              & 90±20        \\
\hline
\end{tabular}
}
\end{table}

\begin{table*}[htbp]
\caption{Performance of different methods}
\label{tab:ovrPerf}
\resizebox{0.98\linewidth}{!}{
\begin{tabular}{lllllllllllll}
\hline
                                Type & Model         & \multicolumn{5}{l}{Historic} & \multicolumn{5}{l}{Recent}         & Avg  \\
                                &               & Rel. & Comp. & Clar. & Acc. & Act. & Rel. & Comp. & Clar. & Acc. & Act. & Ovr. \\ \hline
\multirow{4}{*}{LLMs}    & GPT-4-turbo    & 4.28 & 3.70  & 4.59  & 4.55 & 3.89 & 4.24 & 3.67  & 4.54  & 4.50 & 3.87 & 4.18 \\
                                & GPT-3.5-turbo  & 3.80 & 3.29  & 4.05  & 4.03 & 3.47 & 4.22 & 3.66  & 4.54  & 4.49 & 3.85 & 3.94 \\
                                & Mistral 7B    & 4.23 & 3.51  & 4.70  & 4.60 & 3.90 & 3.97 & 3.24  & 4.44  & 4.37 & 3.60 & 4.05 \\
                                & Llama3 8B     & 4.06 & 3.37  & 4.51  & 4.35 & 3.66 & 3.97 & 3.41  & 4.42  & 4.29 & 3.67 & 3.97 \\ \hline
\multirow{4}{*}{Data Synthesis+SFT} & Self Instruct & 4.29 & 3.77  & 4.79  & 4.51 & 4.09 & 4.25 & 3.74  & 4.75  & 4.46 & 4.06 & 4.27 \\
                                & Evol Instruct & 4.28 & 3.83  & 4.78  & 4.52 & 4.06 & 4.23 & 3.73  & 4.73  & 4.44 & 4.01 & 4.26 \\
                                & Magpie        & 4.02 & 3.73  & 4.52  & 4.21 & 4.05 & 3.97 & 3.65  & 4.45  & 4.16 & 3.96 & 4.07 \\ 
                                & \ourmodel-S        & \underline{4.31} & \underline{3.99}  & \underline{4.83}  & 4.55 & \underline{4.33} & \underline{4.27} & \underline{3.95}  & \underline{4.80}  & 4.52 & \underline{4.29} & \underline{4.38} \\                                 
                                \hline
\multirow{2}{*}{RAG-augmenting-SFT}     & RAFT          & 4.27 & 3.66  & 4.66  & \underline{4.60} & 3.95 & 4.22 & 3.62  & 4.63  & \underline{4.54} & 3.93 & 4.21 \\
                                & DSF           & 4.19 & 3.64  & 4.69  & 4.43 & 3.90 & 4.09 & 3.52  & 4.58  & 4.32 & 3.80 & 4.12 \\ \hline
\makecell[l]{Data Synthesis\\+RAG-augmenting-SFT}       & \ourmodel     & \textbf{4.44} & \textbf{4.16}  & \textbf{4.89}  & \textbf{4.72} & \textbf{4.43} & \textbf{4.37} & \textbf{4.10}  & \textbf{4.86}  & \textbf{4.68} & \textbf{4.40} & \textbf{4.50} \\
\hline
\end{tabular}}
\end{table*}

\textbf{Baseline}. (1) \textit{Proprietary and open-source LLMs} including GPT-4-turbo, GPT-3.5-turbo, Mistral 7B, and Llama3 8B. (2) \textit{Data synthesis+SFT} methods which synthesize instructions and utilize Mistral 7B as the base model for SFT, including Self Instruct~\cite{zhang2023self}, Evol Instruct~\cite{Xu2023WizardLMEL}, and Magpie~\cite{xu2024magpie}. Note that these data-synthesis strategies do not incorporate Retrieval Augmented Generation (RAG) in data synthesis. It is impossible to include retrieved contents as part of the question during the SFT phase. To ensure a fair comparison, we also implement a \textit{variant of the proposed model} \ourmodel-S, which does not use retrieved content in the question. The statistics of generated instructions are shown in Table~\ref{tab:genQAsStatistic}. (3) \textit{RAG-augmenting-SFT} baselines, which utilize Mistral 7B as the base model for retrieval augmented SFT, including RAFT~\cite{zhang2024raft} and DSF~\cite{zhang2024raft}. More details about baselines are discussed in Appendix~\ref{sec:baselinesDetailed}

\textbf{Evaluation}. 
Following previous works~\cite{zhu2023judgelm,zheng2024judging}, we leverage GPT-4-turbo to evaluate the quality of model-generated responses. Specifically, we input the question, the most relevant documents, and the model's response into GPT-4-turbo, prompting it to score the model's answer based on relevance, completeness, clarity, accuracy, and actionability. We further evaluate \ourmodel using DeepSeek-R1 and Llama-3.1-405B, with results provided in the Appendix~\ref{sec:r1result}.

\subsection{Comparative Study}
\label{sec:comp_study}
\textbf{Comparison of Response Quality}. We first compare the performance of 
\ourmodel with various baseline models regarding multi-facet evaluation. The results are shown in Table~\ref{tab:ovrPerf}. We have the following observations.

(1) \ourmodel achieves superior performance in the advertising domain. Compared with proprietary LLMs, \ourmodel achieved improvements of 3.43\%, 12.09\%, 6.69\%, 3.74\%, and 13.69\% over the best-performing GPT-4-turbo in relevance, completeness, clarity, accuracy, and actionability metrics, respectively. It indicates proprietary LLMs only focus on general domain knowledge and do not perform well in the vertical domain, i.e., the advertising domain. Supervised Fine-Tuning (SFT) is necessary for the advertising domain and \ourmodel proposes an efficient data synthesis strategy for SFT.
Compared with other data synthesis strategies, \ourmodel achieved improvements of 3.15\%, 9.98\%, 2.21\%, 4.77\%, and 8.30\% over the best-performing Self Instruct in relevance, completeness, clarity, accuracy, and actionability, respectively. Besides, \ourmodel-S also achieved average improvements of 2.68\% over Self Instruct. This means that our model benefits from conversation-based synthetic data, which enables it to gain insights into users' hidden interests and provide higher-quality responses.

(2) Our model demonstrates significant improvements in Completeness and Actionability, outperforming other baselines by at least 9.13\% and 8.34\%, respectively. These gains are likely due to our use of conversation data, which enhances response quality in two key ways. First, conversation data is highly focused on specific topics, enabling the model to provide more comprehensive and detailed answers, thus improving Completeness. Second, because conversations often explore practical "how-to" details, the model generates more actionable responses, boosting Actionability. 

(3) In our model, RAG-augmenting-SFT has resulted in significant performance improvements. Specifically, \ourmodel demonstrates an overall improvement of 2.73\% compared with \ourmodel-S, which does not use RAG. This indicates that RAG helps the model generate higher-quality responses.

\begin{figure}[!t]
\centering
\includegraphics[width=0.98\linewidth]{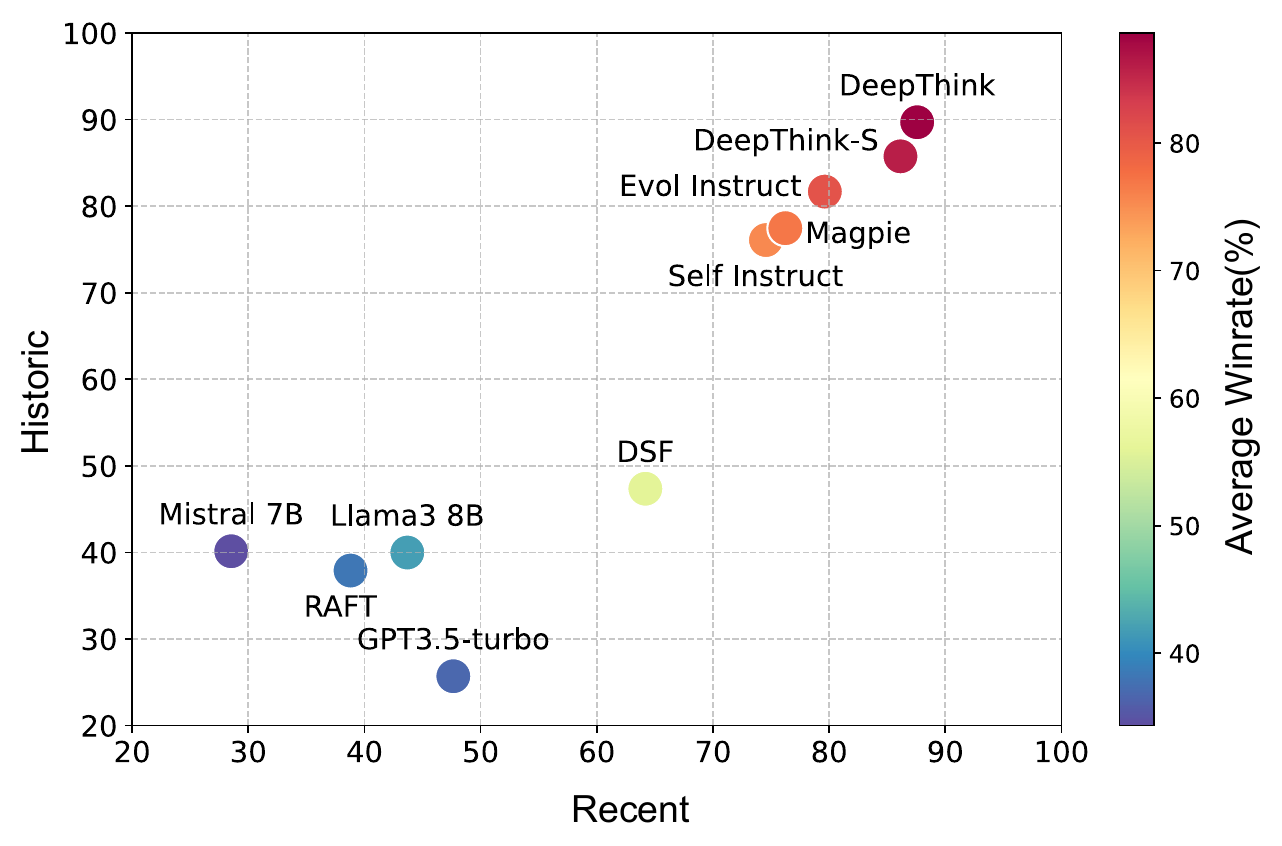}
    \caption{Human Preference Evaluation (WinRate models vs. GPT-4-turbo \%)}
    \label{fig:winrate}
\end{figure}

\textbf{Comparison of Human Preference}. 
We calculate the WinRate of each model in comparison with GPT-4-turbo (the LLM used by the online advertising assistant platform). To reflect the degree of human preference, we use the judgments from GPT-4-turbo. Detailed prompts
are provided in Figure~\ref{fig:evalwinratePrompt}. We report the WinRate on the historical and recent subset in Figure~\ref{fig:winrate}. We also color the baselines by the average win rate. We have the following observations.

(1) Users exhibit a stronger preference for responses generated by \ourmodel. \ourmodel achieves the highest WinRates on both the Historic and Recent datasets, with scores of 89.69\% and 87.58\%, respectively. This indicates that, compared to the original advertising assistant, users prefer the responses from \ourmodel. 

(2) Models fine-tuned by synthesized instructions generally achieve better performance. For example, Self Instruct, Evol Instruct, Magpie, \ourmodel-S, and \ourmodel all achieve a WinRate of at least 70\% on both datasets. 

(3) \ourmodel demonstrates superior performance compared with other instruction synthesis methods for SFT. \ourmodel achieves a 9.90\% improvement over the best-performing baseline, Evol-Instruct. This improvement highlights the effectiveness of our proposed strategy, which leverages conversation data to uncover users' deeper interests and employs an iterative refiner to optimize answers continuously. Our approach not only generates higher-quality answers but also better captures the underlying concerns and interests behind user queries.

\begin{table}[!t]
\caption{Performance of each component in \ourmodel on Recent dataset}
\label{tab:ablation}
\resizebox{0.96\linewidth}{!}{
\begin{tabular}{llllll}
\hline
           & Rel. & Comp. & Clar. & Acc. & Act. \\
           \hline
\ourmodel  & \textbf{4.37} & \textbf{4.10}  & \textbf{4.86}  & \textbf{4.68} & \textbf{4.40} \\
w/o \convaugment     & 4.19 & 3.89  & 4.73  & 4.49 & 4.21 \\
w/o \convrefine     & 4.21 & 3.76  & 4.72  & 4.45 & 4.07 \\
w/o \convaugment, \convrefine & 4.14 & 3.45  & 4.65  & 4.38 & 3.85 \\
\hline
\end{tabular}}
\end{table}

\subsection{Impact of Conversation-base Data Synthesis and Refinement}
We conduct extensive experiments to show the effectiveness of different components in \ourmodel. We conduct a series of ablation studies that involve: (1) removing \convaugmentfull (w/o \convaugment), (2) removing \convrefinefull (w/o \convrefine), and (3) the simultaneous removal of \convaugment and \convrefine in the recent dataset. The results are presented in Table~\ref{tab:ablation}. From these experiments, we draw the following conclusions.

Every component in our model contributes significantly to its performance. When \convaugment is removed, the model exhibits notable declines in response quality: Relevance drops by 4.12\%, Completeness by 5.12\%, Clarity by 2.67\%, Accuracy by 4.06\%, and Actionability by 4.32\%. Similarly, removing \convrefine results in reductions of 3.72\% in Relevance, 8.34\% in Completeness, 2.83\% in Clarity, 4.88\% in Accuracy, and 7.52\% in Actionability. Furthermore, when both \convaugment and \convrefine are removed simultaneously, the model's performance degrades even more significantly, with Relevance decreasing by 5.32\%, Completeness by 15.90\%, Clarity by 4.27\%, Accuracy by 6.37\%, and Actionability by 12.52\%. These results clearly demonstrate the importance and effectiveness of each component in our model.

(2) The removal of \convrefine has the most significant impact on the model's performance. This demonstrates the critical role of \convrefine in leveraging conversational context to enhance response quality. Specifically, \convrefine refines responses by utilizing assessment feedback to filter out irrelevant or meaningless information from dialogues, thereby significantly improving the overall quality of the generated answers.

(3) \convaugment has a particularly strong impact on improving the relevance of the model's responses. When \convaugment is removed, the relevance of the model's answers shows the most significant decline, dropping by 4.17\%. This is likely because \convaugment generates a broader range of high-quality instructions by simulating real-world user conversations. By closely mimicking how users naturally communicate, \convaugment ensures that the generated instructions are more aligned with actual user queries, thereby enhancing the relevance of the model's responses.


\begin{figure}[!t]
\centering
\includegraphics[width=0.98\linewidth]{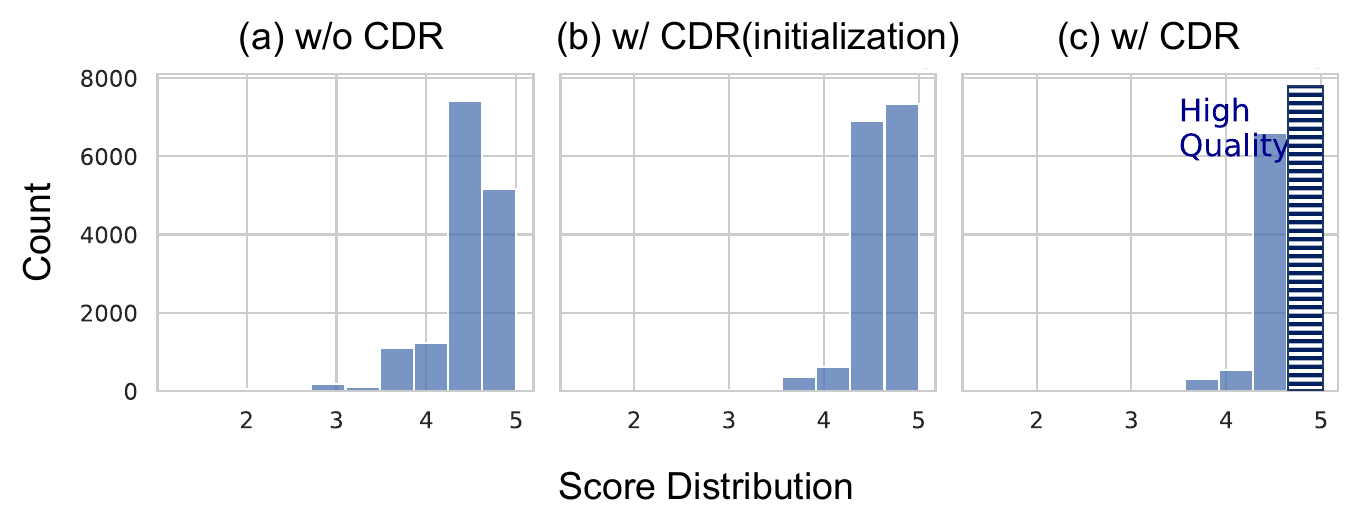}
    \caption{Score distribution of the instructions}
    \label{fig:scoreDist}
\end{figure}

Furthermore, we conduct ablation studies on the refiner in \convrefinefull. Specifically, we implement the three variants: (1) the original answers obtained by synthesized conversation without refinement (w/o \convrefine), (2) answers initially refined by only the conversation contexts (w/ \convrefine initialization), and (3) answers iteratively refined by assessment feedback (w/ \convrefine). We report the distribution of assessment scores in Figure~\ref{fig:scoreDist}, and we make the following observations.

(1) The initial effect of the refiner is significant. The average score has been increased from $4.63$ to $4.75$ after refiner initialization, representing an improvement of 2.59\%. This indicates that refinement leveraging conversational context enhances response quality.

(2) Feedback-guided refiner further increases the ratio of high-quality answers, i.e., with a score of five. This demonstrates that the assessment feedback effectively guides the refiner to fine-tune responses, making them better aligned with user preferences.

\begin{table}[!t]
\caption{Performance of Imitation-based and Synthesis-only seed data}
\label{tab:imqVsSynQ}
\resizebox{1\linewidth}{!}{
\begin{tabular}{cccccccc}
\hline
                & Sim. &Rel. &Comp. &Clar. &Acc. &Act.\\
\hline
Synthesis-only  & 0.76 & 4.14 & 3.42  & 4.63 & \textbf{4.37} & 3.81 \\
Imitation-based & \textbf{0.79} & \textbf{4.15} & \textbf{3.46}  & \textbf{4.64}  & \textbf{4.37} & \textbf{3.84}
\\
\hline
\end{tabular}}
\end{table}
\subsection{Necessary of Imitation}
We analyze the differences between instructions generated by GPT-4-turbo using two distinct approaches: imitation-based, which replicates the style of real user questions, and synthesis-only, i.e., \textit{SELF-QA}, which generates instructions without such imitation. Utilizing the all-mpnet-base-v2 model, we obtain embeddings for each instruction and for real user questions from the Recent evaluation dataset. In addition to the five evaluation dimensions, we also calculate the similarity(Sim.) between the centroid of the imitation-based instruction embeddings and the centroid of real user question embeddings, as well as between the centroid of synthesis-only instruction embeddings and the centroid of real user question embeddings.

As shown in Table~\ref{tab:imqVsSynQ}, imitation-based instructions exhibit higher similarity (Sim.=0.79) to actual user questions compared with synthesis-only instructions (Sim.=0.76). Additionally, models fine-tuned on imitation-based data demonstrate improved performance across various metrics, including Relevance, Completeness, Clarity, Accuracy, and Actionability, compared with those trained with synthesis-only data. Specifically, imitation-based methods achieve improvements of 0.24\% in Relevance, 1.17\% in Completeness, and 0.22\% in Clarity. These results indicate that imitation-based instruction data more closely align with real user queries, leading to enhanced model performance.

\subsection{Performance of RAG-augmenting-SFT }
\label{sec:rag_in_sft}

\begin{figure}[htbp]
\centering
\includegraphics[width=0.96\columnwidth]{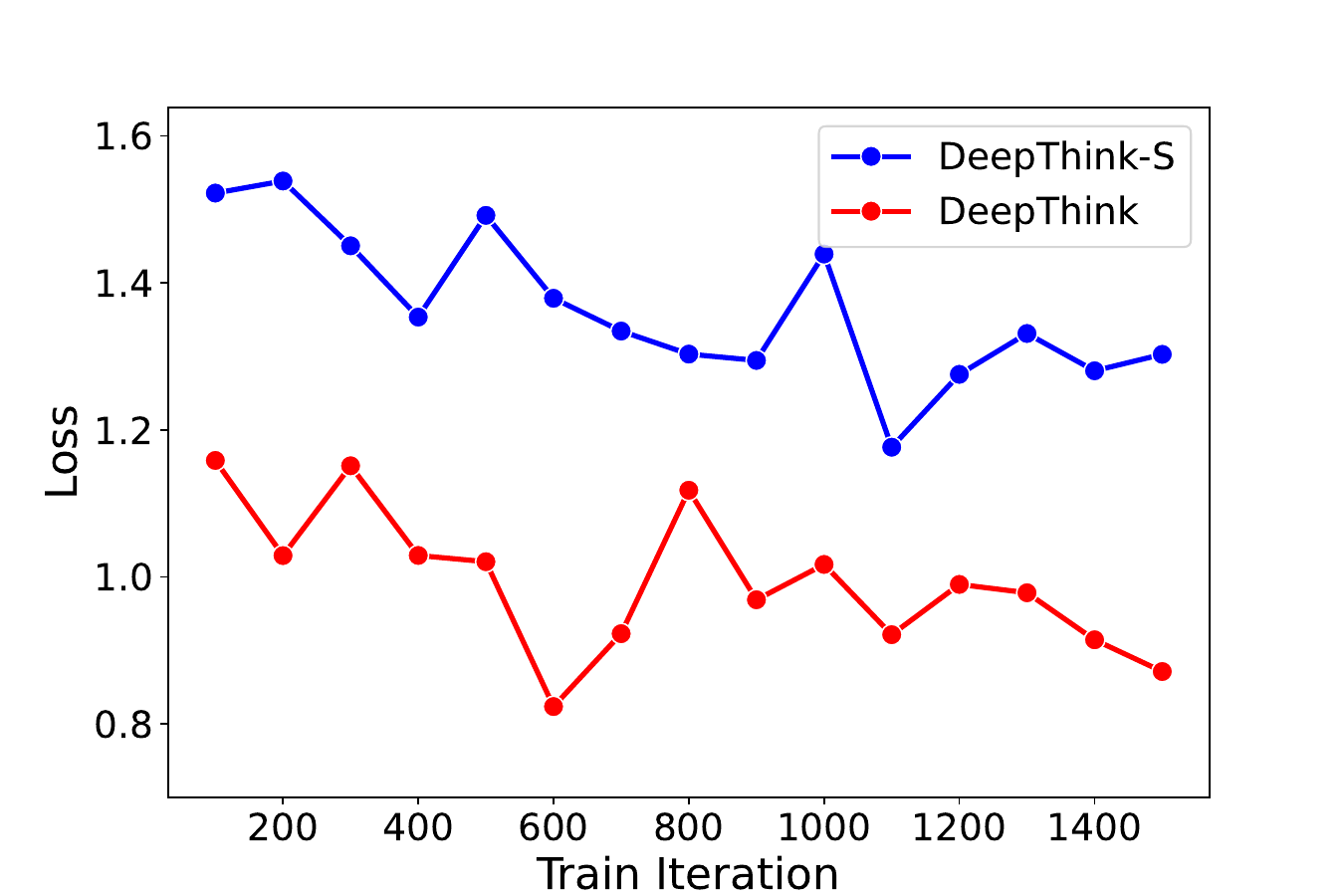}
    \caption{Training loss trend of \ourmodel with and without RAG-augmenting-SFT on Recent}
    \label{fig:trainloss}
\end{figure}

To evaluate the impact of Retrieval-Augmented Generation (RAG) on Supervised Fine-Tuning, we first compare the loss of \ourmodel-S  (i.e., uses only the original questions without any retrieved documents) with \ourmodel (i.e., uses retrieved documents). As shown in Figure~\ref{fig:trainloss}, the training loss for \ourmodel-S is significantly higher than that for \ourmodel, with an average increase of 37.28\%. This discrepancy can be attributed to the distinct reliance on knowledge sources during the Supervised Fine-Tuning phase. Specifically, solely relying on the knowledge poses a non-trivial challenge for LLMs. In contrast, integrating retrieved documents within the instructional contexts allows learning objectives to align effectively with more accurate responses, reinforcing LLM's ability during the SFT phase.

\begin{figure}[htbp]
\centering
\includegraphics[width=0.96\linewidth]{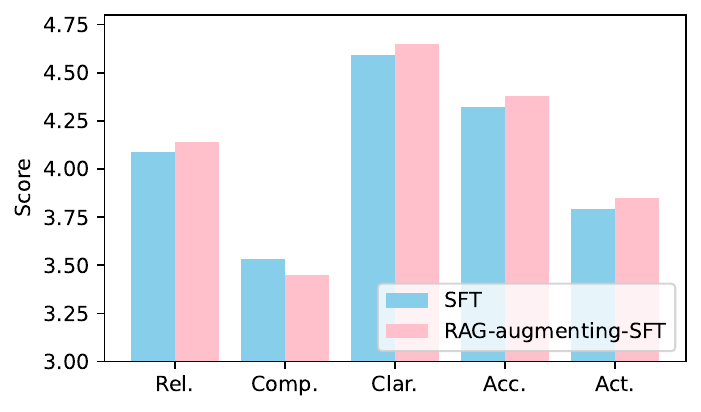}
    \caption{Performance of SFT and RAG-augmenting-SFT on Recent}
    \label{fig:ragVsnorag}
\end{figure}

We have shown that \ourmodel achieves a notable improvement regarding all performance metrics compared with \ourmodel-S in Table~\ref{tab:ovrPerf}. We further remove the conversation component and implement (1) SFT that uses only seed instructions for fine-tuning and (2) RAG-augmenting-SFT that uses seed instructions along with retrieved documents. As shown in Figure~\ref{fig:ragVsnorag}, incorporating relevant documents as part of the input on lower-quality instructions also helps the model better understand contextual relationships and enhances QA capabilities.

%% file: conclustion.tex
\section{Conclusion}
In this paper, we propose \ourmodel, a novel framework designed to improve the performance of large language models (LLMs) in domain-specific question-answering tasks. By integrating three key components: data synthesis based on conversations, data refinement based on conversations, and supervised fine-tuning (SFT) enhanced with retrieval, \ourmodel addresses the critical challenge of adapting LLM to understand and meet hidden user needs in vertical domains. Our experiments demonstrate that \ourmodel outperforms GPT-4-turbo+RAG by 7.92\% across the evaluation metrics. 

\section{Limitations}
This study has several limitations. First, the experimental validation was exclusively conducted within the advertising domain, which may constrain the generalizability of our methodology to other vertical domains (e.g., e-commerce, education, or healthcare). Future research should extend the evaluation framework by conducting cross-domain experiments to verify the robustness of our approach. Second, the assessment protocol relied primarily on GPT-4-turbo, DeepSeek-R1, and Llama-3.1-405B for automated evaluation, potentially introducing model-specific biases. Future work can explore(1) implementing human-in-the-loop evaluation with advertising professionals to assess practical utility, and (2) incorporating real-world A/B testing with actual advertisers to measure performance metrics in production environments.

%% file: appendix.tex
\appendix
\newpage
\section{Implementation}
\label{sec:imp}
In data synthesis, we sample 15 seed queries, and the maximum number of conversation turns is three. We iterate for three rounds to refine the answers.

We utilize the Mistral 7B Instruct model~\cite{jiang2023mistral} as the base model for fine-tuning. In the training phase, following prior works~\cite{taori2023stanford,Wang2022SelfInstructAL,Xu2023WizardLMEL}, we apply supervision on the output tokens' loss. The fine tuning is performed using the Xtuner framework~\cite{2023xtuner} with a learning rate $lr=2e-5$, a warm-up ratio of 0.03, and a batch size of 1. We employ the LORA training method with hyper-parameters rank $r$ set to 64, $\alpha$ set to 16, and dropout rate $p$ set to 0.05. During the generation phase, text generation is performed using vLLM~\cite{kwon2023efficient} with a temperature coefficient $T=0.7$.
In the context of Retrieval-Augmented Generation (RAG), we utilize the LangChain framework to process domain-specific data. We employ CharacterTextSplitter to segment the data into text chunks with a chunk size of 512 and an overlap of 32. These chunks are then embedded using the pre-trained all-mpnet-base-v2 model\footnote{https://huggingface.co/sentence-transformers/all-mpnet-base-v2}, and the embeddings are stored in a Chroma database. During retrieval, we calculate the similarity between the question and the stored chunks, selecting the top 3 most similar chunks as the retrieval results.

\section{Baselines}
\label{sec:baselinesDetailed}
We compare \ourmodel with the following instruction synthesis baselines.
\begin{itemize}
    \item Self Instruct~\cite{zhang2023self}: a method which leverages a small set of seed data and a pretrained language model to synthesize a large amount of instructional data for fine-tuning.
    \item Evol Instruct~\cite{Xu2023WizardLMEL}: a method that starts with a basic set of instructions and employs a large language model to iteratively rewrite them, progressively enhancing their complexity. This approach generates a wide array of instructional data with varying levels of complexity. 
    \item Magpie~\cite{xu2024magpie}: a self-synthesis method that leverages the autoregressive feature of aligned LLMs like Llama-3-Instruct to auto-generate 4 million high-quality instructions, with 300K selected for fine-tuning.
\end{itemize}
We also compare \ourmodel with the following baselines that use RAG to augment SFT. 
\begin{itemize}
    \item RAFT~\cite{zhang2024raft}: a training method that enhances large language models (LLMs) for open-book question answering by utilizing Chain-of-Thought (CoT) during the Supervised Fine-Tuning (SFT) phase. It incorporates both relevant and irrelevant documents in the context, training the model to ignore the irrelevant ones and focus on citing useful information in its output. 
    \item DSF: performing standard supervised finetuning, without documents in context. We follow the same setting as mentioned in RAFT.
\end{itemize}
\textbf{Remarks.} The goal of RAFT is to train the model to distinguish which documents are relevant to the question so that the model can answer based on these documents. On the other hand, \ourmodel aims to help the model identify knowledge in the documents that is not only relevant to the question but also aligns with the user's intent (since the answers after CDR incorporate conversational information, uncovering the deep user intent in the question). We aim for this process to be implicit, avoiding the reliance on explicit CoT, which can sometimes be inaccurate. User intentions are complex and diverse, and inappropriate or stereotypical CoT reasoning may hinder the model's ability to fully capture the user's true intent~\cite{turpin2023language}. Besides, unlike RAFT, we did not deliberately introduce irrelevant documents in instructions that could confuse the model.

\section{Comparison of Response Quality Evaluated by Different LLMs}
\label{sec:r1result}
To further validate the effectiveness of \ourmodel, we conducted an additional evaluation using a slow-thinking reasoning model. Specifically, we employed DeepSeek-R1~\cite{guo2025deepseek}\footnote{We locally deployed the open-source DeepSeek-R1-Distill-Qwen-32B model and DeepSeek-R1-Distill-Llama-70B model} and Llama-3.1-405B to assess the performance of representative baseline methods and \ourmodel on the Recent dataset, following the same evaluation prompt template described in Section~\ref{sec:comp_study}. As demonstrated in Table~\ref{tab:r1comp}, the experimental results reveal that when evaluated through the slow-thinking reasoning framework of DeepSeek-R1, \ourmodel achieves consistent conclusions with those obtained from GPT-4-turbo. This alignment persists across multiple evaluation dimensions, suggesting that our method maintains robust performance even under more deliberate and systematic reasoning paradigms and different LLM-based evaluators.

\begin{table}[htbp]
\caption{Performance of each component in \ourmodel on Recent dataset evaluated by Different LLM Evaluators.}
\label{tab:r1comp}
\resizebox{0.98\linewidth}{!}{
\begin{tabular}{lllllll}
\hline
          Evaluator & Setting & Rel. & Comp. & Clar. & Acc. & Act. \\
           \hline
           \multirow{5}{*}{Llama-3.1-405B} &GPT-4-turbo & 4.50 & 3.79 & 4.60 & 4.52 &3.94 \\
           &Self Instruct & 4.67 & 4.01 & 4.78 & 4.62 & 4.17 \\
           &Evol Instruct & 4.63 & 3.99 & 4.77 & 4.63 &4.17 \\
           &RAFT & 4.60 & 3.81 & 4.63 & 4.59 & 4.05 \\ &\ourmodel  & \textbf{4.78} & \textbf{4.25}  & \textbf{4.79}  & \textbf{4.73} & \textbf{4.57} \\
          
           \hline
            \multirow{5}{*}{\shortstack{DeepSeek-R1-\\Distill-Qwen-32B}} &GPT-4-turbo & 4.43 & 3.83 & 4.55 & 4.56 &4.14 \\
           &Self Instruct & 4.53 & 3.92 & 4.72 & 4.57 & 4.27 \\
           &Evol Instruct & 4.48 & 3.86 & 4.70 & 4.52 &4.27 \\
           &RAFT & 4.60 & 3.92 & 4.70 & 4.72 & 4.31 \\
            & \ourmodel  & \textbf{4.75} & \textbf{4.26}  & \textbf{4.82}  & \textbf{4.77} & \textbf{4.65}

           \\ \hline
           \multirow{5}{*}{\shortstack{DeepSeek-R1-\\Distill-Llama-70B}} &GPT-4-turbo & 4.40 & 3.71 & 4.67 & 4.61 &4.07 \\
           &Self Instruct & 4.42 & 3.79 & 4.84 & 4.55 & 4.23 \\
           &Evol Instruct & 4.36 & 3.72 & 4.83 & 4.49 &4.22 \\
           &RAFT & 4.48 & 3.74 & 4.78 & 4.68 & 4.22 \\
                  
&\ourmodel  & \textbf{4.60} & \textbf{4.06}  & \textbf{4.86}  & \textbf{4.70} & \textbf{4.56} \\
\hline
\end{tabular}}
\end{table}

\section{Comparison of Synthetic Data Quality}
We sample 1000 questions each from three synthetic datasets: \ourmodel, Self Instruct, and Evol Instruct, as well as from the Recent evaluation dataset. Using the all-mpnet-base-v2 model, we obtain embeddings for these questions. We then use t-SNE to assess the distribution similarity between \ourmodel, Self Instruct, Evol Instruct, and Recent datasets. Furthermore, we calculate the centroid embedding for each dataset and assessed the similarity of the centroid embeddings between \ourmodel, Self Instruct, Evol Instruct, and Recent. This approach allows us to evaluate which synthetic data generation method yields data that is more comparable to real user questions.

\begin{figure}[htbp]
\centering
\includegraphics[width=1\linewidth]{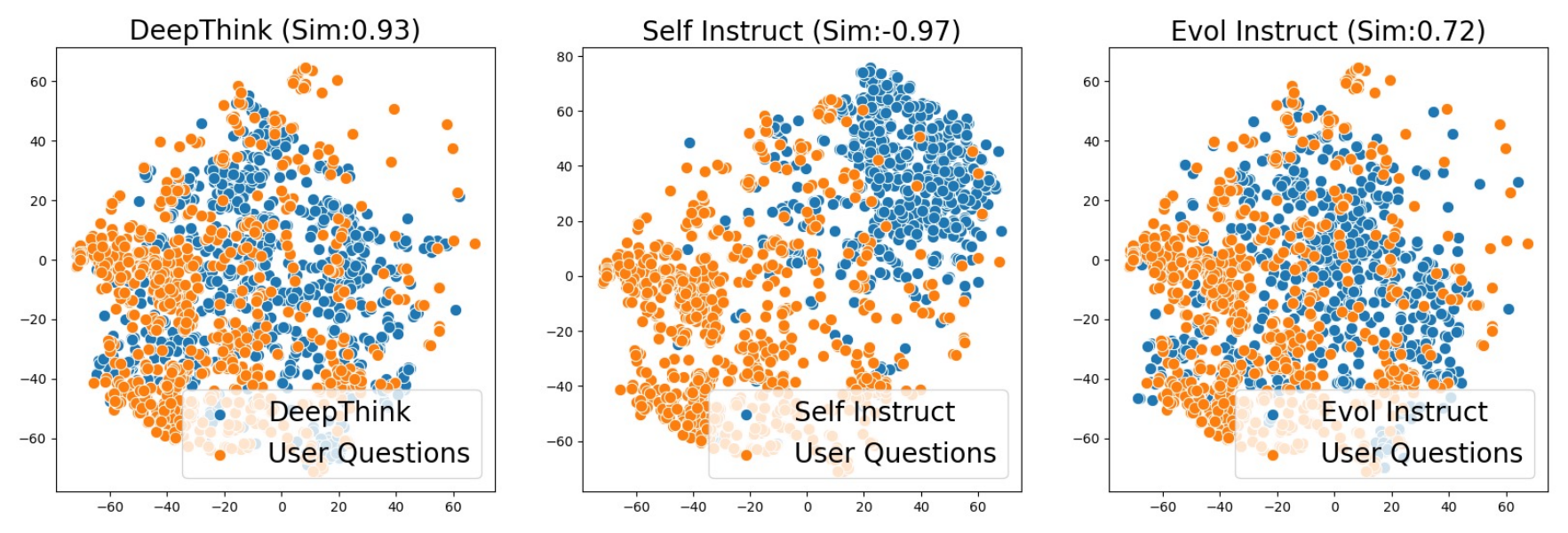}
    \caption{Similarity between different synthetic data methods and real user questions}
    \label{fig:qualComp}
\end{figure}

\begin{figure}[htbp]
\centering
\includegraphics[width=0.85\linewidth]{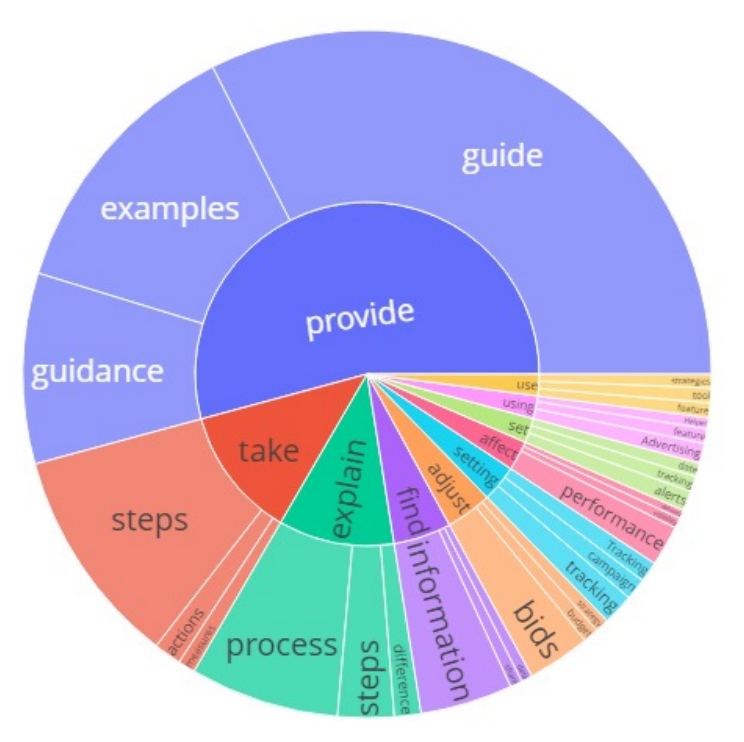}
    \caption{Top 10 Most Common Root Verbs (Inner) and Their Top 3 Direct Noun Objects (Outer) in \ourmodel}
    \label{fig:deepthinkPie}
\end{figure}

\begin{figure}[htbp]
\centering
\includegraphics[width=0.85\linewidth]{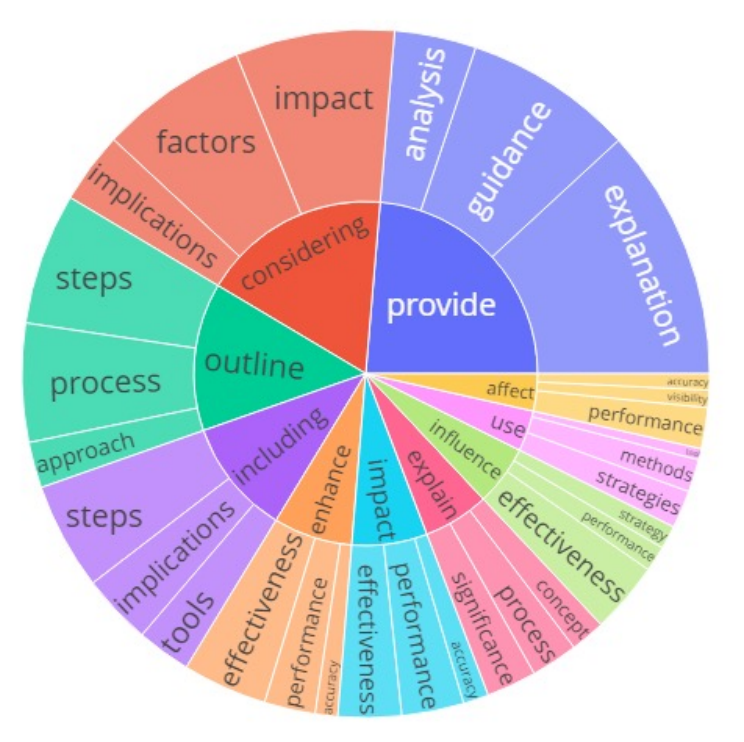}
    \caption{Top 10 Most Common Root Verbs (Inner) and Their Top 3 Direct Noun Objects (Outer) in Evol Instruct}
    \label{fig:evolPie}
\end{figure}

\begin{figure}[htbp]
\centering
\includegraphics[width=0.85\linewidth]{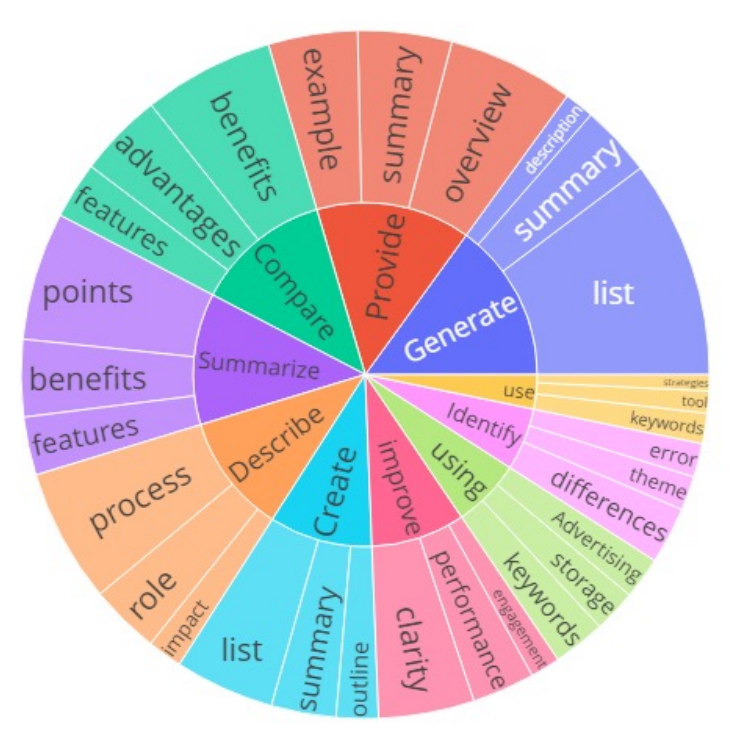}
    \caption{Top 10 Most Common Root Verbs (Inner) and Their Top 3 Direct Noun Objects (Outer) in Self Instruct}
    \label{fig:selfPie}
\end{figure}

As depicted in Figure~\ref{fig:qualComp}, a comparison between \ourmodel and other methods such as Self Instruct and Evol Instruct reveals that the instruction data generated by \ourmodel exhibits significantly higher relevance to actual user questions. The centroid smilarity of \ourmodel is 0.93, while Self Instruct got -0.97 and Evol Instruct got 0.72. This marked relevance demonstrates that \ourmodel, through its data construction approach that simulates conversational formats and style as found on real advertising platforms, generates instructions that not only better reflect user expression but also satisfy the actual demands users may present in specific scenarios.

Furthermore, we follow the previous work~\cite{xu2024magpie} and show the visualization of root verbs and their direct noun objects. Figure ~\ref{fig:deepthinkPie},~\ref{fig:evolPie} and~\ref{fig:selfPie} visualize the top common root verbs and their direct noun objects of DeepThink,  Evol Instruct and Self Instruct dataset, respectively.
A notable finding is that in \ourmodel, the verb "provide" holds a significantly larger proportion compared to other synthesis approaches. Additionally, expressions such as "-guidance" and "-example" are types of questions that users are more inclined to ask in the advertising domain. This result further validates that \ourmodel can generate more questions that users would actually ask in this field.

\section{Full Parameters vs. LoRA Finetuning}
We conduct two types of fine-tuning, full-parameter fine-tuning and LoRA fine-tuning, on Self Instruct, Evol Instruct, and our proposed model, \ourmodel. Specifically, we employ QLoRA~\cite{dettmers2024qlora}, a quantization-based efficient finetuning improvement of LoRA.These are subsequently evaluated on the Recent dataset. For full-parameter fine-tuning, we employ the Mistral 7B base model, while for LoRA fine-tuning, the Mistral 7B Instruct is selected as the foundation model. Our evaluation focus primarily on the relevance of the model's responses, as this metric is a crucial indicator of the model's accuracy and utility in understanding and generating answers. Relevance of the responses is critical because it directly influences the model’s capability to solve problems, authenticity, and user satisfaction. As depicted in Figure, we observe that full-parameter fine-tuning significantly underperformed compared to LoRA fine-tuning. One possible reason for this discrepancy is the divergence between the synthetic training data and the distribution of real user questions, which hampers the model's ability to generalize to authentic user data in full-parameter tuning. Additionally, our \ourmodel displays superior performance in full-parameter fine-tuning compared to Self Instruct and Evol Instruct, which partially demonstrates the effectiveness of our imitation-based instruction synthesis method, as it yields instructions that more closely align with the distribution of real user questions.

\begin{figure}[htbp]
\centering
\includegraphics[width=0.7\linewidth]{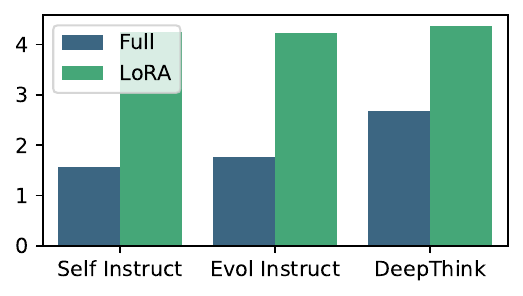}
    \caption{Performance between Full parameters and QLoRA finetuning}
    \label{fig:fullandlora}
\end{figure}

\section{Case Study}

\subsection{Synthesized Conversations and Refining Responses}
\label{sec:scrr}
\begin{figure*}[htbp]
\centering
\includegraphics[width=1\linewidth]{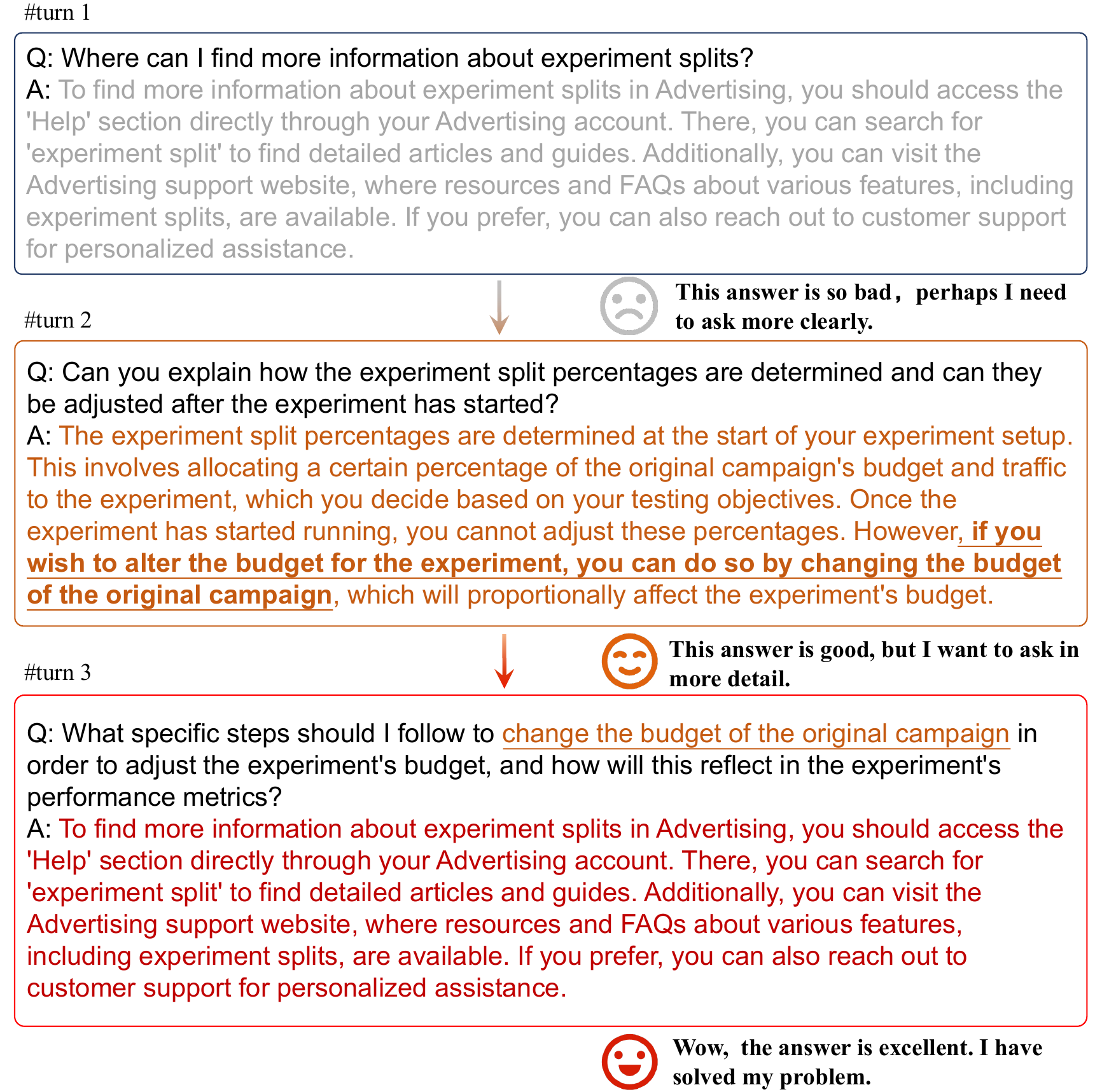}
    \caption{The case of conversation between the user and the assistant synthesized by $\ourmodel$.}
    \label{fig:goodCDS}
\end{figure*}

\begin{figure*}[htbp]
\centering
\includegraphics[width=1\linewidth]{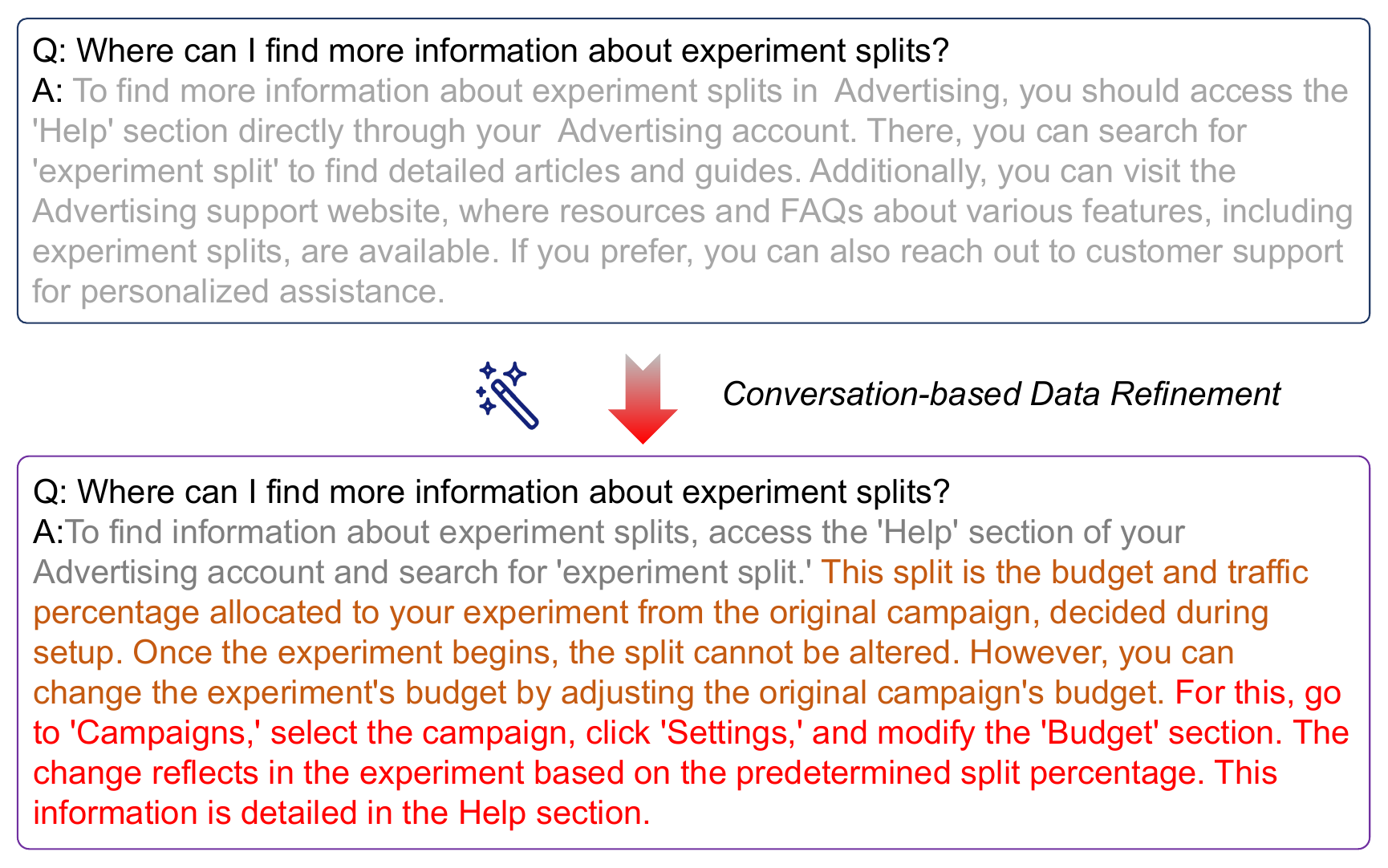}
    \caption{The case of the response after Conversation-based Data Refinement.}
    \label{fig:goodCDR}
\end{figure*}

We present a conversation synthesized by DeepThink, as shown in Figure~\ref{fig:goodCDS}. In this conversation, the user initially asks a rather vague question, leading to a less helpful and broad response. In the next round, the user clarifies the question, receiving a more precise answer. Finally, the user follows up on a specific detail in the response, obtaining a deeper and more detailed answer. This type of conversation closely mirrors the scenarios that people encounter when using language models, as described in Section~\ref{sec:intro}, and $\ourmodel$ successfully synthesizes such conversations. Furthermore, Figure~\ref{fig:goodCDR} shows that $\ourmodel$ improves the original answer to the question using Conversation-based Data refinement.

\subsection{Online platform vs. DeepThink}
\begin{figure*}[htbp]
\centering
\includegraphics[width=1\linewidth]{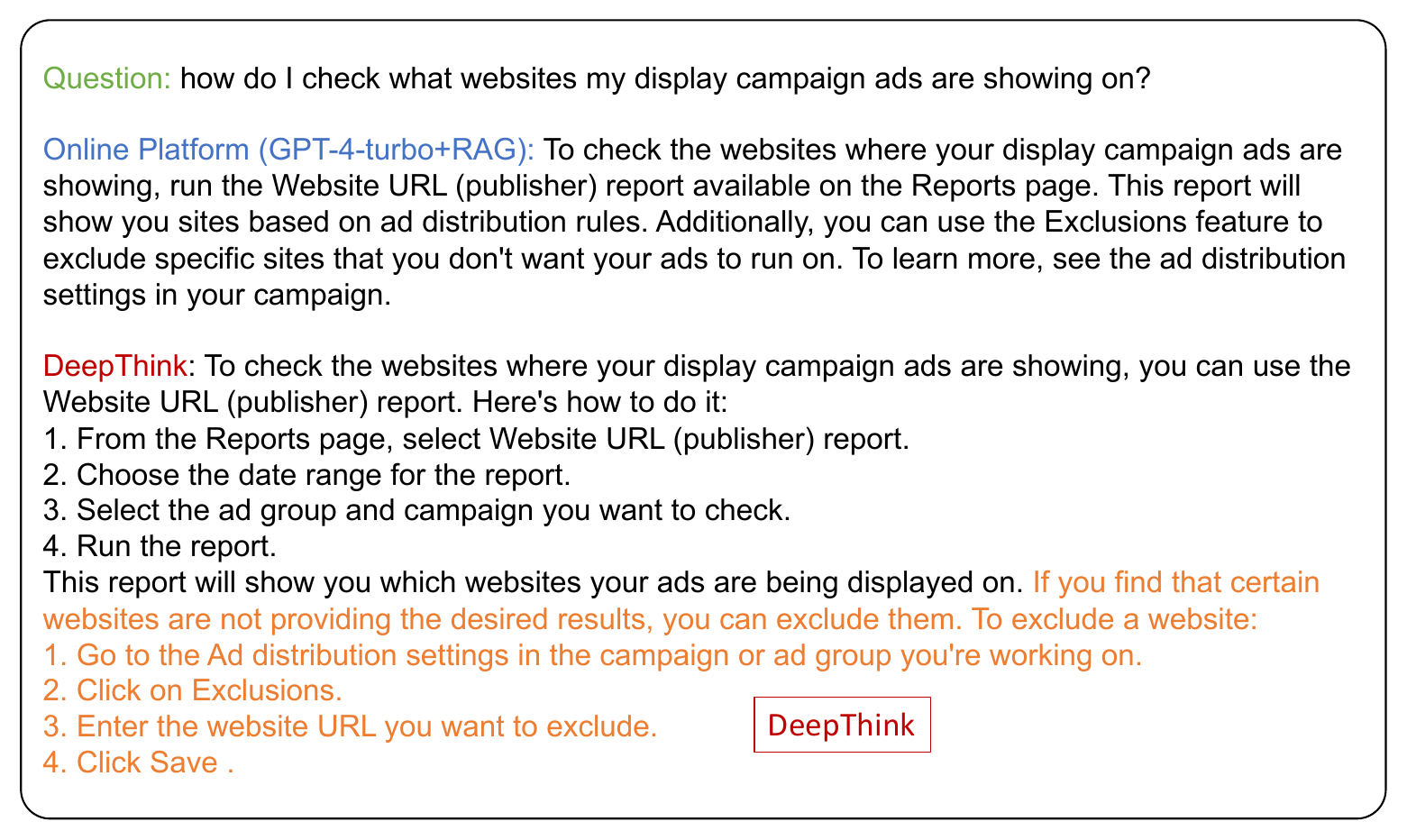}
    \caption{Online platform answer vs. DeepThink(Case 1)}
    \label{fig:case1}
\end{figure*}

\begin{figure*}[htbp]
\centering
\includegraphics[width=1\linewidth]{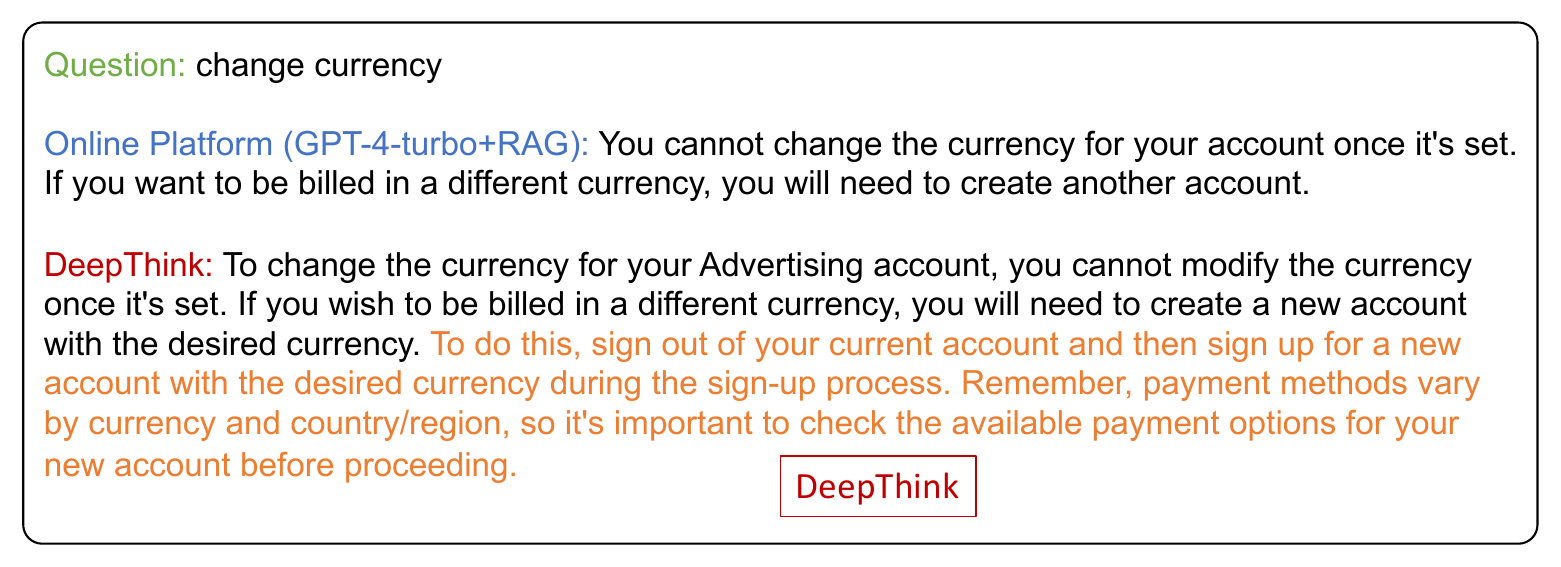}
    \caption{Online platform answer vs. DeepThink(Case 2)}
    \label{fig:case2}
\end{figure*}

We demonstrate the answers for the online advertising platform(GPT-4-turbo+RAG) and $\ourmodel$ regarding the same real user questions. They are shown in Figure~\ref{fig:case1} and~\ref{fig:case2}. The cases show answers generated by $\ourmodel$ better than the online advertising platform in completeness and actionability.

\section{Prompts}
\label{sec:appPrompts}
\begin{figure*}[htbp]
\centering
\includegraphics[width=1\linewidth]{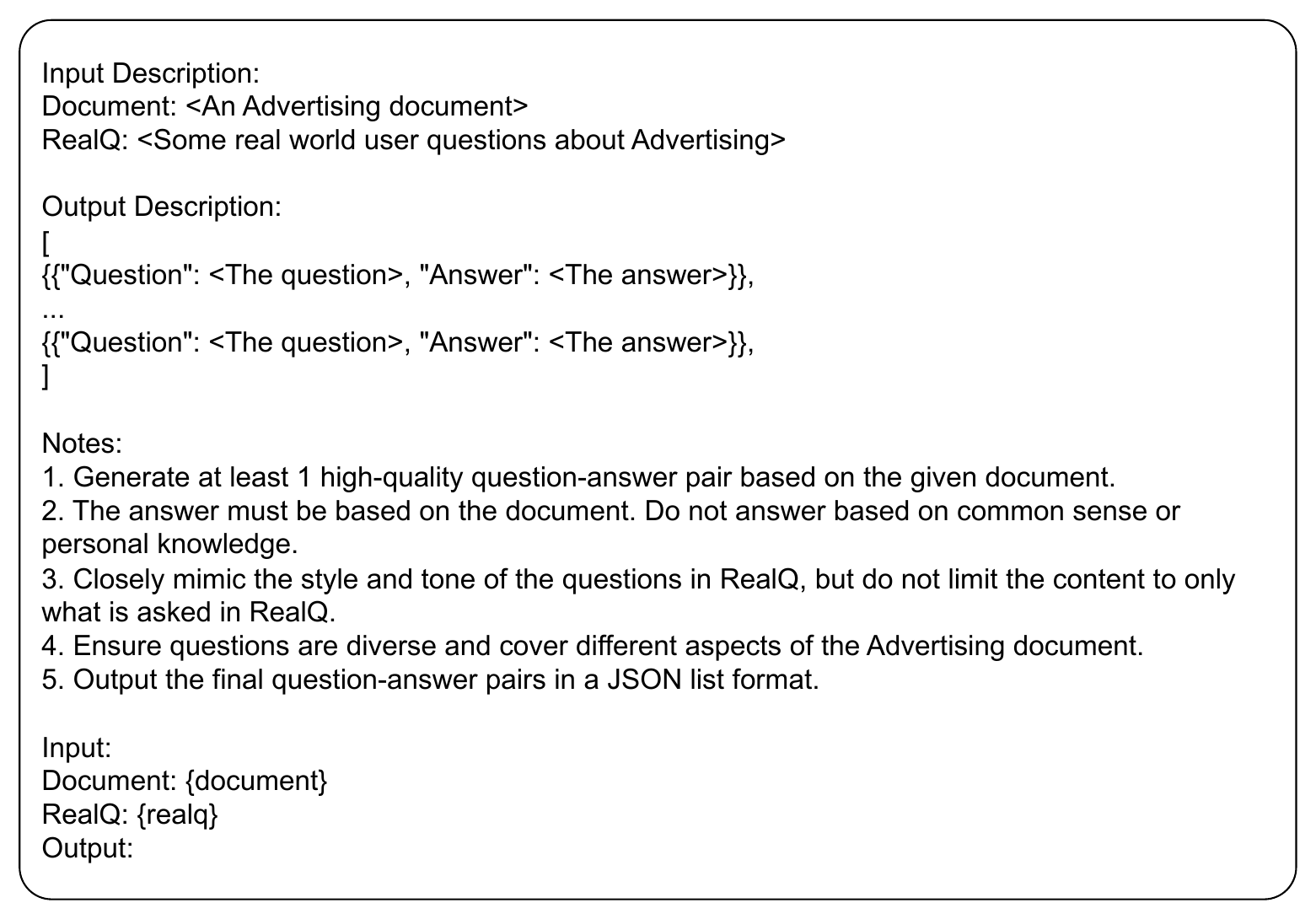}
    \caption{The prompt of extracting Seed QAs from documents}
    \label{fig:docQAsPrompts}
\end{figure*}

\begin{figure*}[htbp]
\centering
\includegraphics[width=1\linewidth]{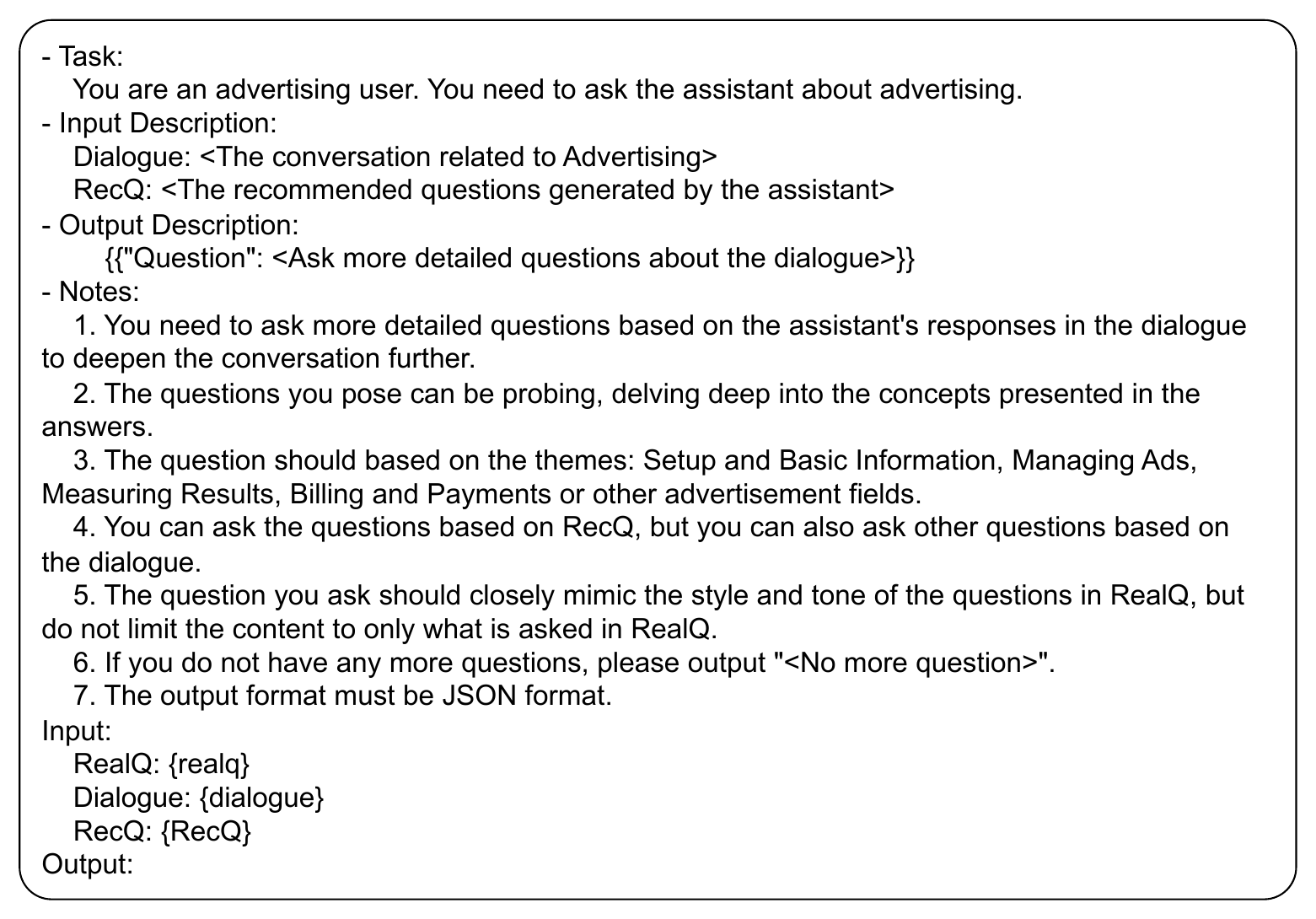}
    \caption{The prompt of $\Inquirer$ in Conversation-based Data Synthesis}
    \label{fig:userAgentPrompt}
\end{figure*}

\begin{figure*}[htbp]
\centering
\includegraphics[width=1\linewidth]{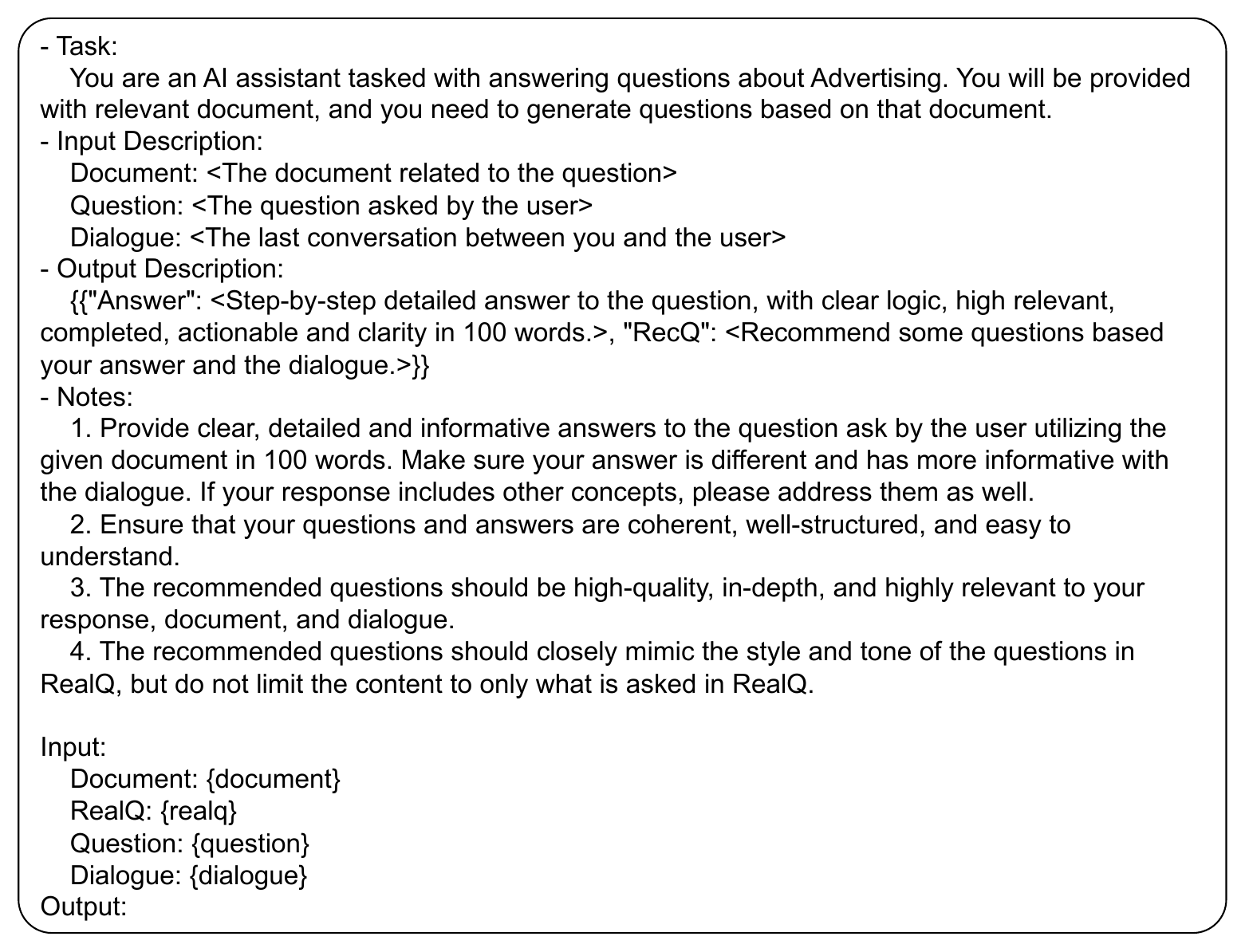}
    \caption{The prompt of $\Assistant$ in Conversation-based Data Synthesis}
    \label{fig:assitantAgentPrompt}
\end{figure*}

\begin{figure*}[htbp]
\centering
\includegraphics[width=1\linewidth]{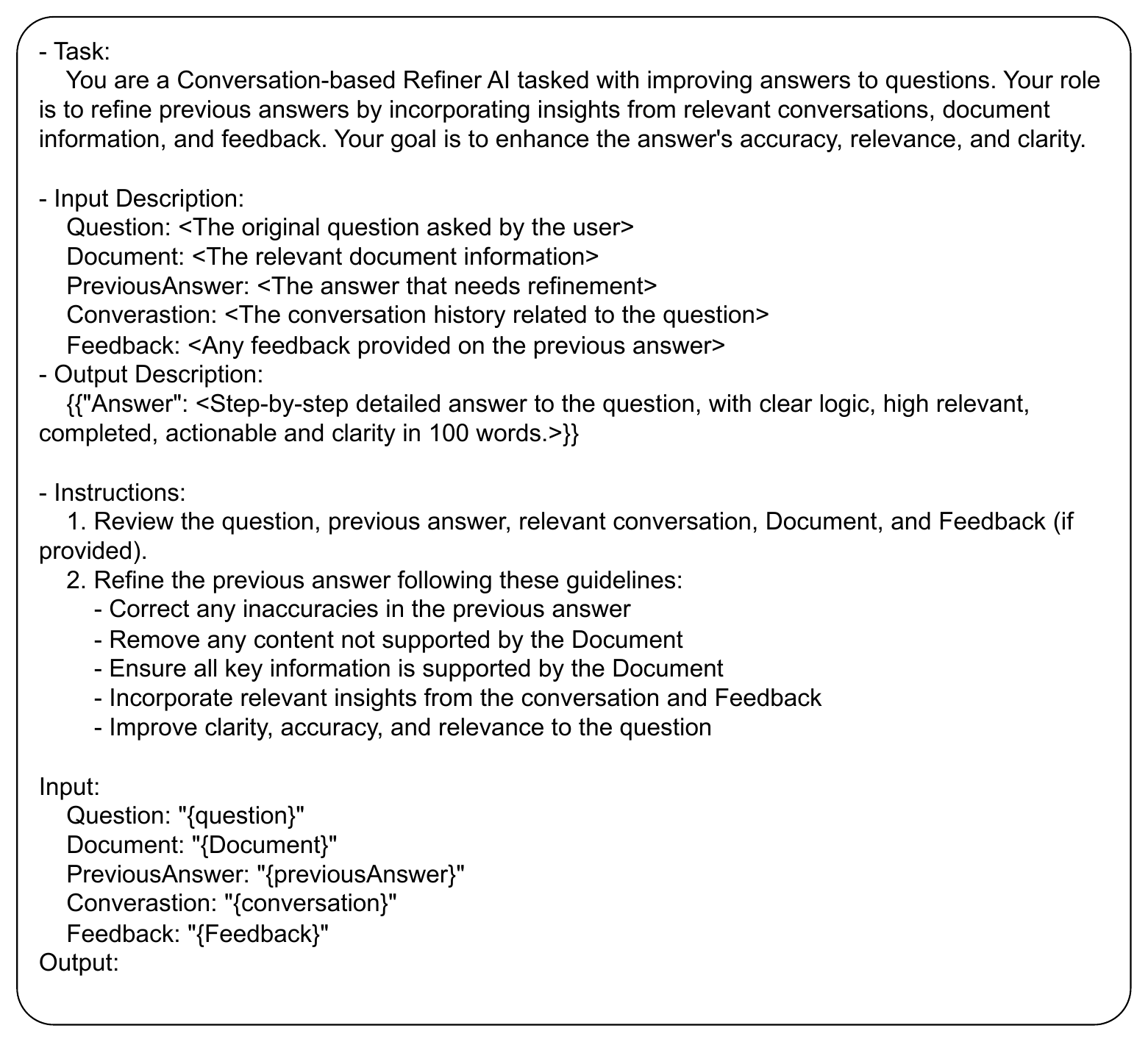}
    \caption{The prompt of Conversation-based Data Refinement}
    \label{fig:refinerPrompt}
\end{figure*}

\begin{figure*}[htbp]
\centering
\includegraphics[width=1\linewidth]{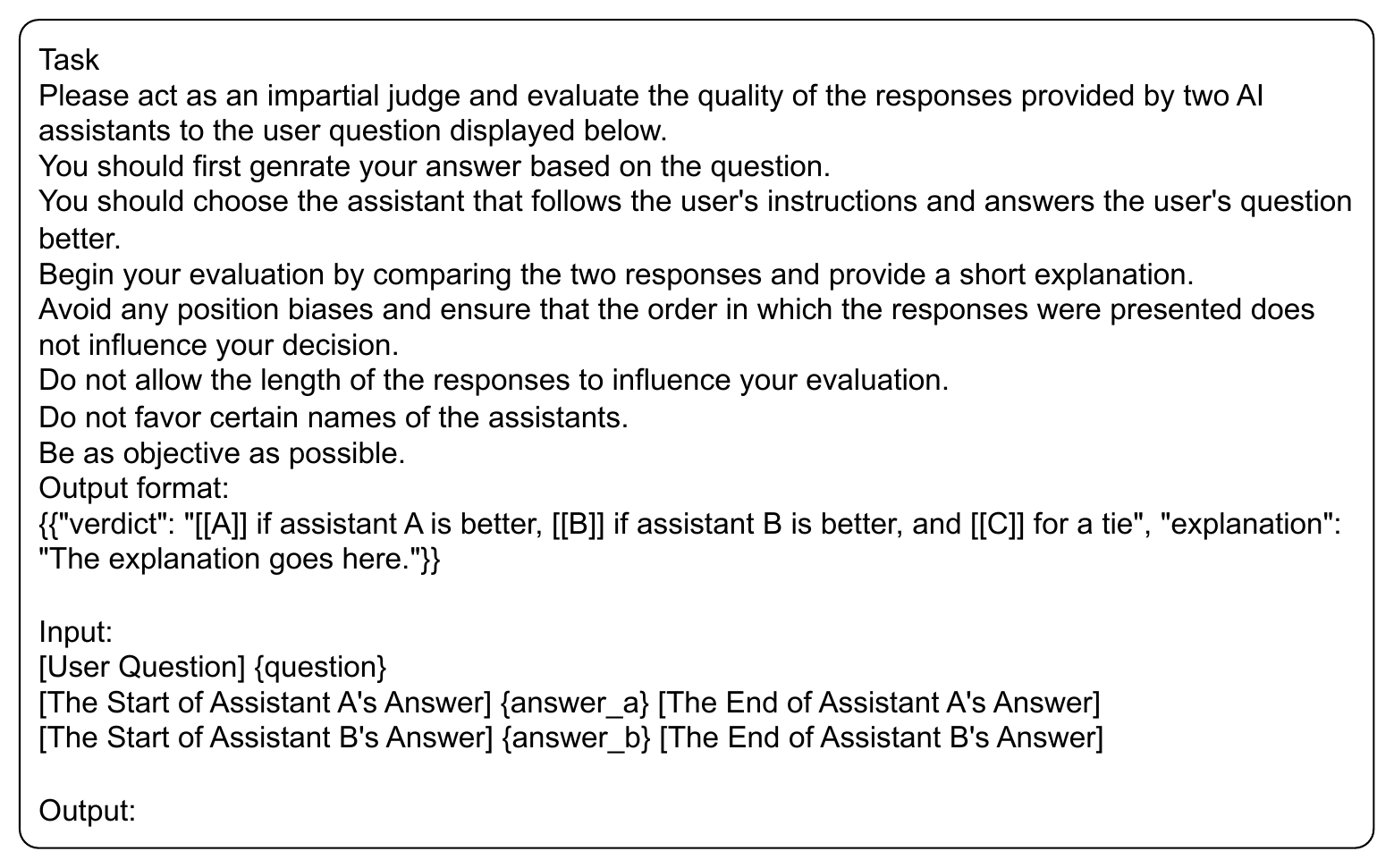}
    \caption{The prompt of the evaluation prompt based on the relevant document}
    \label{fig:evalDocPrompt}
\end{figure*}

\begin{figure*}[htbp]
\centering
\includegraphics[width=1\linewidth]{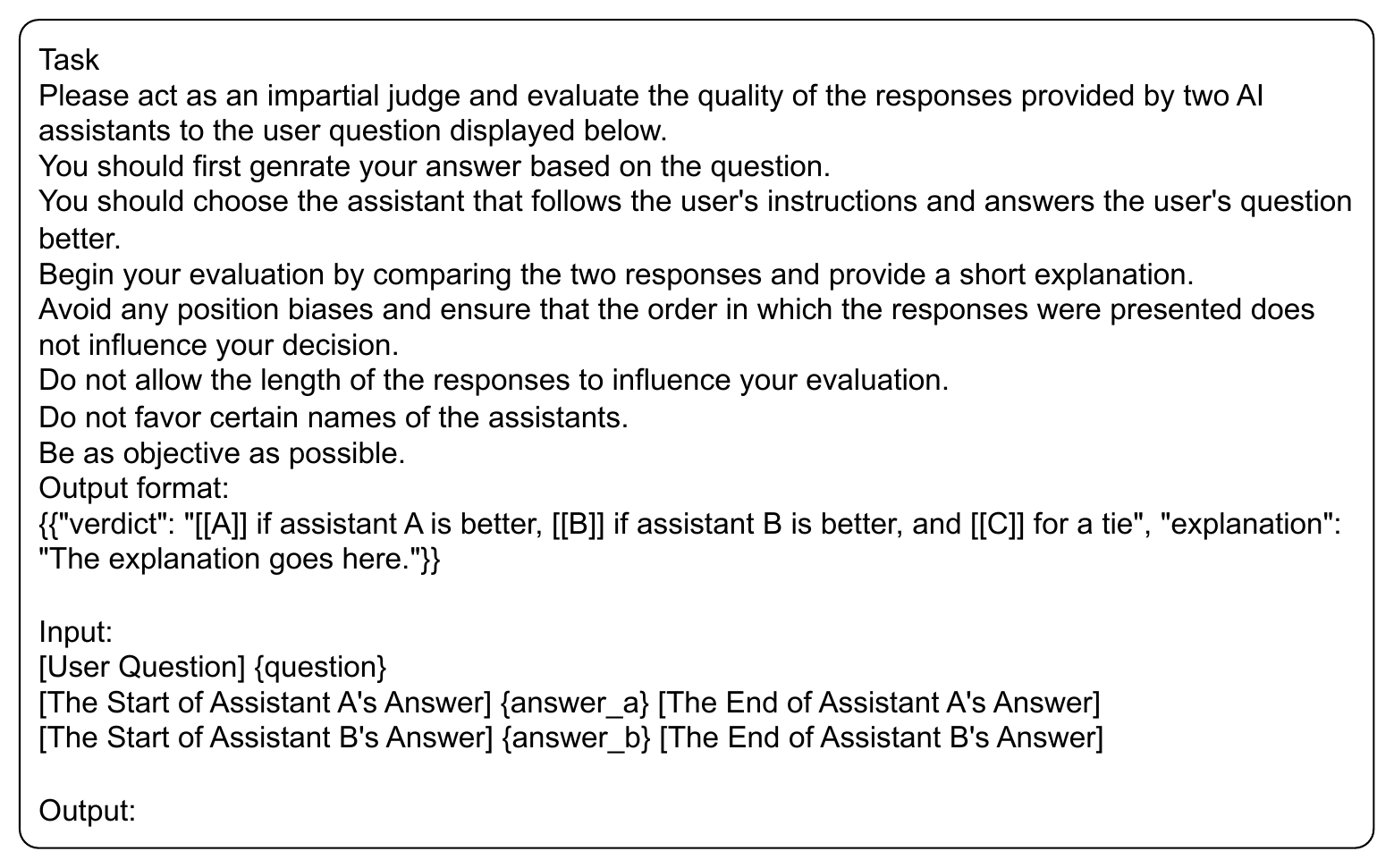}
    \caption{The prompt of the winrate evaluation prompt}
    \label{fig:evalwinratePrompt}
\end{figure*}
